\pdfoutput=1
\documentclass[accepted]{uai2021} 

\usepackage[british]{babel}

\usepackage{natbib} 
\usepackage{microtype}
\usepackage{booktabs} 
\usepackage{amsfonts}       
\usepackage{amsmath}
\usepackage{bm}
\usepackage{dsfont} 
\usepackage{url}            
\usepackage{verbatim}
\usepackage{wrapfig}
\usepackage{tikz}
\usetikzlibrary{bayesnet}
\usetikzlibrary{arrows}

\usepackage{graphicx}
\usepackage{caption}
\usepackage[tableposition=above]{caption} 
\usepackage{subcaption}
\usepackage{pinlabel}

\usepackage{xr-hyper}

\usepackage{hyperref}

\usepackage{sidecap}
\sidecaptionvpos{figure}{c}

\usepackage{amsmath,amsfonts,bm}

\def\eqref#1{equation~\ref{#1}}

\def\1{\bm{1}}

\DeclareMathAlphabet{\mathsfit}{\encodingdefault}{\sfdefault}{m}{sl}
\SetMathAlphabet{\mathsfit}{bold}{\encodingdefault}{\sfdefault}{bx}{n}

\newcommand{\E}{\mathbb{E}}

\newcommand{\R}{\mathbb{R}}
\newcommand{\Z}{\mathbb{Z}}

\newcommand{\KL}{D_{\mathrm{KL}}}

\DeclareMathOperator{\id}{\mathds{1}}

\DeclareMathOperator*{\argmin}{arg\,min}

\usepackage[utf8]{inputenc} 
\usepackage[T1]{fontenc}    
\usepackage{hyperref}       
\usepackage{url}            
\usepackage{nicefrac}       
\usepackage{comment}        
\usepackage{cleveref}       

\usepackage{algorithm}
\usepackage{algorithmic}
\usepackage[makeroom]{cancel}
\usepackage{enumitem}
\usepackage{amsthm}
\usepackage{amssymb}
\usepackage{xcolor}
\usepackage{dsfont} 

\usepackage{caption}
\usepackage[tableposition=above]{caption} 
\usepackage{subcaption}

\title{Hierarchical Indian Buffet Neural Networks for Bayesian Continual Learning}

\author[1]{\href{mailto:<skessler@robots.ox.ac.uk>?Subject=Your UAI 2021 paper}{Samuel Kessler}{}}
\author[2]{Vu Nguyen}
\author[1]{Stefan Zohren}
\author[1]{Stephen J. Roberts}

\affil[1]{%
    University of Oxford
}
\affil[2]{%
    Amazon Adelaide
}
  
\begin{document}
\maketitle

\begin{abstract}
 We place an Indian Buffet process (IBP) prior over the structure of a Bayesian Neural Network (BNN), thus allowing the complexity of the BNN to increase and decrease automatically. We further extend this model such that the prior on the structure of each hidden layer is shared globally across all layers, using a Hierarchical-IBP (H-IBP). We apply this model to the problem of resource allocation in Continual Learning (CL) where new tasks occur and the network requires extra resources. Our model uses online variational inference with reparameterisation of the Bernoulli and Beta distributions, which constitute the IBP and H-IBP priors. As we automatically learn the number of weights in each layer of the BNN, overfitting and underfitting problems are largely overcome. We show empirically that our approach offers a competitive edge over existing methods in CL.
\end{abstract}

\section{Introduction}
\label{sec:intro}

Humans have the ability to continually learn, consolidate their knowledge and leverage previous experiences when learning a new set of skills. In Continual Learning (CL) an agent must also learn continually, presenting several challenges including learning online, avoiding forgetting and efficiently allocating resources for learning new tasks. In CL, a neural network model is required to learn a series of tasks, one by one, and remember how to perform each. The model is given a set of $M$ tasks sequentially $\mathcal{T}_t$ for $t = 1, ... \, M$. Where each task is comprised of a dataset. 
The model will lose access to the training dataset for task $\mathcal{T}_t$ but will be continually evaluated on the test sets for all previous tasks $\mathcal{T}_i$ for $i \leq t$, we will introduce the problem setting more formally in the next section.

The principal challenges to CL are threefold, firstly models need to overcome \textit{catastrophic forgetting} of old tasks; a neural network will exhibit forgetting of previous tasks after having learnt a few tasks \citep{Goodfellow2015}. Secondly, models need to leverage knowledge transfer from previously learnt tasks for learning a new task $\mathcal{T}_t$. And finally, the model needs to have enough neural resources available to learn a new task and adapt to the complexity of the task at hand.

One of the main approaches to CL involves the use of the natural sequential learning approach embedded within Bayesian inference. The prior for task $\mathcal{T}_i$ is the posterior which is obtained from the previous task $\mathcal{T}_{i-1}$. This enables knowledge transfer and offers an approach to overcome catastrophic forgetting. Previous Bayesian CL approaches have leveraged Laplace approximations \citep{Kirkpatrick} and variational inference \citep{VCL} to aid computational tractability. Whilst Bayesian methods solve the first and second objectives above, the third objective of ensuring that the BNN has enough neural resources to adapt its complexity to the task at hand is not necessarily achieved. For instance, additional neural resources can alter performance on MNIST classification (see Table 1 in \citep{Blundell}). This is a problem as the amount of neural resources required for a current task, may not be enough (or may be redundant) for a future task. Propagating a poor approximate posterior from one task will alter performance for all subsequent tasks.

Non-Bayesian neural networks use additional neurons to learn new tasks and prevent overwriting previous knowledge thus overcoming forgetting. The neural networks which have been trained on previous tasks are frozen and a new neural network is appended to the existing network for learning a new task \citep{Rusu}. The problem with this approach is that of scalability: the number of neural resources increases linearly with the number of tasks. The scalability issue has been tackled with selective retraining and expansion with a group regulariser \citep{Yoon}. However this solution is unable to shrink and so are vulnerable to overfitting if misspecified when starting CL. Moreover knowledge transfer and prevention of catastrophic forgetting are not solved in a principled manner, unlike approaches couched in a Bayesian framework.

As the resources required are typically unknown in advance, we propose a BNN which adds or withdraws neural resources automatically in response to the data. This is achieved by drawing on Bayesian nonparametrics to learn the structure of each hidden layer of a BNN. Thus, the model size adapts to the amount of data seen and the difficulty of the task. This is achieved by using a binary latent matrix $Z$, distributed according to an Indian Buffet Process (IBP) prior \citep{Griffiths2011}. The IBP prior on an infinite binary matrix, $Z$, allows inference on which and how many neurons are required for each data point in a task. The weights of the BNN are treated as draws from non-interacting Gaussians \citep{Blundell}. Catastrophic forgetting is overcome by repeated application of the Bayesian update rule, embedded within variational inference \citep{VCL}. We summarise the contributions as follows. We present a novel BNN using an IBP prior and its hierarchical extension to automatically learn the complexity of each hidden layer according to the task difficulty. The model's effective use of resources is shown to be useful in CL. We derive a variational inference algorithm for learning the posterior distribution of the proposed models. In addition, our model elegantly bridges two separate CL approaches: expansion methods and Bayesian methods (more commonly referred to as regularization based methods in CL literature).

\looseness=-1

\section{Indian Buffet Neural Networks}

\label{sec:model}
We introduce the CL problem setting in Section~\ref{sec:cl}, variational Bayesian approaches to CL in Section~\ref{sec:bcl}. We present the IBP prior in Section~\ref{sec:ibp} and the IBP prior on the latent binary matrix $Z$ is then applied to a BNN such that the complexity of each hidden layer can be learnt from the data in Section~\ref{sec:ibp_bnn}. In Section~\ref{sec:hibp}, the Hierarchical IBP prior (H-IBP) is introduced and applied to the BNN to encourage a more regular structure. Thus, the use of an IBP and H-IBP prior over the hidden states of the BNN can be readily used together with the Bayesian CL framework presented, and so automatically adapt its complexity according to the task.

\looseness=-1

\subsection{Continual Learning}
\label{sec:cl}
\emph{Continual learning} (CL) is a setting whereby a model must learn a set of tasks sequentially, while maintaining performance across all tasks. In CL, the model is shown a set of $M$ tasks sequentially $\mathcal{T}_{t}$ for $t= 1, \ldots M$. 
Each task is comprised of a dataset such that $\mathcal{T}_t: \mathcal{D}_t = \left \{ (x_{i}, y_{i}) \right  \} $ for $i = 1, \ldots, N_t$. The inputs $x_i \in \R^d$ and outputs can be $y_i \in \R$ in the case of regression or a categorical variable for classification. Although the model will lose access to the training dataset for task $\mathcal{T}_t$, it will be continually evaluated on all previous tasks $\mathcal{T}_i$ for $i \leq t$. $t$ can be used as a task identifier informing the agent when to start training on a new task or what task to being tested. For a comprehensive review of CL scenarios see \citet{VanDeVen, Hsu2018}.

\looseness=-1

\subsection{Bayesian Continual Learning}

\label{sec:bcl}
The CL process can be decomposed into Bayesian updates where the approximate posterior for $\mathcal{T}_{t-1}$ can be used as a prior for task $\mathcal{T}_{t}$. Variational CL (VCL) \citep{VCL} uses a BNN to perform the prediction tasks where the network weights are independent Gaussians. The variational posterior from previous tasks is used as a prior for new tasks. Consider learning the first task $\mathcal{T}_1$, and $\bm{\phi}$ are the variational random variables, then the variational posterior is $q_1(\bm{\phi}|\mathcal{D}_1)$. For the subsequent task, access to $\mathcal{D}_1$ is lost and the prior will be $q_1(\bm{\phi}|\mathcal{D}_1)$, optimization of the ELBO will yield the variational posterior $q_2(\bm{\phi}|\mathcal{D}_2)$. Generalising, the negative ELBO for the $t$-th task is:
\begin{equation}
\begin{aligned}
    \mathcal{L}(\bm{\phi}, \mathcal{D}_t) &= \KL \left[q_t(\bm{\phi})||q_{t-1}(\bm{\phi}|\mathcal{D}_{t-1}) \right] \\
    & \quad - \E_{q_t}[\log p(\mathcal{D}_t | \bm{\phi})].
\end{aligned}
\end{equation}
The first term acts to regularise the posterior such that it is close to previous task's posterior and the second term is the log-likelihood of the data for the current task.

\looseness=-1

\subsection{Indian Buffet Process prior}
\label{sec:ibp}
Matrix decomposition aims to represent the data $X$ as a combination of latent features: $X = Z A + \epsilon$ where $X \in \R^{N \times D}$, $Z \in \Z_{2}^{N \times K}$, $A \in \R^{K \times D}$ and $\epsilon$ is an observation noise. Each element in $Z$ corresponds to the presence or absence of a latent feature from $A$. Specifically, $z_{ik} = 1$ corresponds to the presence of a latent feature $A_{k}$ in observation $X_{i}$ and $k \in \{1, \cdots, \infty\}$ all columns in $Z$ with $k>K$ are assumed to be zero. In a scenario where the number of latent features $K$ is to be inferred, then the IBP prior on $Z$ is suitable \citep{Doshi-Velez2009}.\footnote{We provide a notebook to demonstrate how the IBP prior can be used for the matrix factorization \url{ https://bit.ly/3asylU5}. In particular, we illustrate how one doesn't need to specify the number of latent dictionary items to infer. This means we do not need to set the hidden state size of a BNN for our model.}

One representation of the IBP prior is the stick-breaking formulation \citep{teh07a}. The probability $\pi_k$ is assigned to the column $z_k$ for $k \in \{1, \cdots, \infty \}$, whether a feature has been selected is determined by $z_{nk} \sim \textrm{Bern}(\pi_k)$. This parameter $\pi_k$ is generated according to the following stick-breaking process: $v_k \sim \textrm{Beta}(\alpha, 1)$, and $\pi_k = \prod^{k}_{i=1} v_i$, thus $\pi_k$ decreases exponentially with $k$. The Beta concentration parameter $\alpha$ controls how many features one expects to see in the data, the larger $\alpha$ is, the more latent features are present.

\looseness=-1

\subsection{Adaptation with the IBP prior}
\label{sec:ibp_bnn}
Consider a BNN with $k_j$ neurons for each layer $j \in \{1, ... \, J\}$ layers. Thence, for an arbitrary activation $f$, the binary matrix $Z$ is applied elementwise $h_{j} = f(h_{j-1} W_j) \circ Z_j$ where $h_{j-1} \in \R^{N \times k_{j-1}}$, $W_j \in \R^{k_{j-1} \times k_j}$, $Z_j \in \Z_{2}^{N \times k_j}$, and where $\circ$ is the elementwise product and $N$ is the number of data points per batch. We have ignored biases for simplicity. $Z_j$ is distributed according to an IBP prior. The IBP prior has some suitable properties for this application: the number of neurons sampled grows with $N$ and the promotion of ``rich get richer" scheme for neuron selection \citep{Griffiths2011}. For convenience, we term the IBP BNN as IBNN for the remainder of the paper.

The number of neurons selected grow or contract according to the variational objective; which depends on the complexity of the data. This allows for efficient use of neural resources which is crucial to a successful CL model. The variational objectives for the IBP prior and BNN are introduced further down the line in Section~\ref{sec:ibp_svi} and Section~\ref{sec:hibp_svi}. Additionally, the ``rich get richer" scheme is useful since the common neurons are selected across tasks enabling knowledge transfer and preventing forgetting.

As a standard practice in variational inference with a Bayesian nonparametric prior, we use a truncation level $K$, to the maximum number of features in the variational IBP posterior. \citep{doshi2009variational} present bounds on the marginal distribution of $X$ in a matrix factorisation setting and show that the bound decreases exponentially as $K$ increases. A similar behaviour is expected for our application.

\looseness=-1

\subsection{Structured VI}
\label{sec:ibp_svi}
Structured stochastic VI (SSVI) has been shown to perform better inference of the IBP posterior than mean-field VI in deep latent variable models \citep{Singh}. Hence, this inference method has been chosen for learning and presented next. 

A separate binary matrix $Z_j$ can be applied to each layer $j \in \{1, ...\, J\}$ of a BNN. The subscript $j$ is dropped for clarity. The structured variational approximation is: $q(\phi) = \prod^{K}_{k=1}q(v_k)q(\bm{w}_{k})\prod^{N}_{i=1}q(z_{ik}|v_k)$, where the random variables are $\phi = \{v_k, \bm{w}_k, Z_k\}$ and the variational parameters $\varphi=\{ \alpha_k, \beta_k, \bm{\mu}_k, \bm{\sigma}_k\}$ for all $k$. The variational distributions over $\phi$ are defined below and the variational IBP posterior is truncated to $K$. The constituent distributions are $q(v_k) = \textrm{Beta}(\alpha_k, \beta_k)$, $\pi_k = \prod^{k}_{i=1}v_i$,  $q(z_{ik}) = \textrm{Bern}(\pi_{k})$ and the BNN weights are independent draws from $q(\bm{w}_k) = \mathcal{N}(\bm{\mu}_k, \bm{\sigma}^2_k \id)$. Having defined the structured variational objective, the negative ELBO is:
\looseness=-1
\begin{figure}
     \centering
     \raisebox{0.6em}{
        \scalebox{1.3}{
         \begin{tikzpicture}
          \node[latent]                            (z) {$z_{ik}$};
          \node[latent, above=of z]                (v) {$v_k$};
          \edge {v} {z} ; %
          \plate {z} {(z)} {$N$} ;
          \plate [inner sep=.25cm] {} {(v)(z)} {$K$} ;
        \end{tikzpicture}
        }
    }
    \hspace{2.0em}
    \scalebox{1.3}{
     \begin{tikzpicture}
      \node[latent]                            (z) {$z_{ijk}$};
      \node[latent, above=of z]                (pi) {$\pi_{jk}$};
      \node[latent, right=of pi]                (v) {$v_k$};
      \edge {pi} {z} ; %
      \edge {v} {pi} ; %
    
      \plate {z} {(z)} {$N$} ;
      \plate [inner sep=.25cm] {} {(pi)(z)} {$J$} ;
      \tikzset{plate caption/.append style={below=2.8 of v.south east}}
      \plate [inner sep=0.40cm, xshift=-0.2cm, yshift=0.15cm] {} {(v)(pi)(z)} {$K$} ;
    \end{tikzpicture}
    }
     \caption{The graphical model for the structured variational posterior approximation for \textbf{Left}, the IBP and \textbf{Right}, the H-IBP\protect\footnotemark. Using the language of the eponymous IBP prior metaphor, $k$ (dishes) indicates the neurons, the number selected $K$ adjusts flexibly. $j$ (restaurants) is the number of layers which is fixed. $i$ (customers) is a data point.} 
\label{fig:graphical_model_ibp_hibp}
\end{figure}
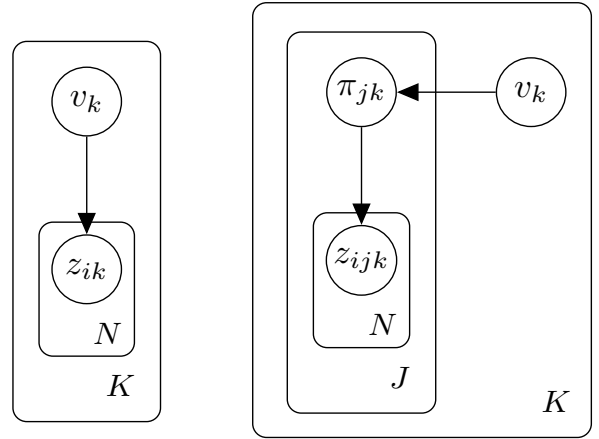
\looseness=-1

\begin{equation}
\begin{aligned}
\mathcal{L}(\phi, \mathcal{D}) &= \KL(q(\bm{v})||p(\bm{v}))+ \KL(q(\bm{w})||p(\bm{w})) \\
&\quad - \sum^{N}_{i = 1} \E_{q(\phi)}\left[\log p(y_i | \mathbf{x}_i, \mathbf{z}_{i}, \bm{w})\right] \\
&\quad + \sum^{N}_{i = 1}\KL(q(\mathbf{z}_{i}|\bm{\pi})|| p(\mathbf{z}_{i}|\bm{\pi})).
\end{aligned}
\end{equation}

We note that the Bernoulli random variables have an explicit dependence on the stick-breaking probabilities in the structured variational approximation. However, this explicit dependence between parameters is removed in a mean field approximation. The variational parameter Beta parameters will control the expected hidden state size for the BNN and is automatically inferred. For sequential Bayesian updates the posterior is then used as the next task's prior, thus a prior only needs to be designed for the first task. 



\looseness=-1

\section{Hierarchical IBNN}
\label{sec:hibp}
In the previous section, we presented the IBNN model to allow a BNN to automatically select the number of neurons for each layer according to the data. For a multi-layer BNN, one can apply the IBP prior independently for each layer. However, we can add an inductive bias to ensure that the inferred number of neurons are similar for across all layers of an MLP and ensure that information is shared across layers. To do this, we propose a hierarchical IBP prior \citep{thibaux2007hierarchical} for neuron selection across multiple layers. The number of neurons from all layers are generated from the same global prior, thus will discourage irregular structure in the BNN (a BNN with adjacent wide and narrow layers might be inferred when using independent priors on each hidden layer). Of course, this property might not be desirable for all use cases, however the majority of BNNs used in the literature have a regular structure. We term our model as HIBNN and present the graphical model which describes the hierarchical IBP prior in Figure~\ref{fig:graphical_model_ibp_hibp}. This model has the advantage of having fewer learnable parameters as neuron selection will be driven by the global prior in comparison to having a single IBP per layer.

\looseness=-1

\subsection{Adaptation with the Hierarchical IBP}
The global probability of selecting the neuron positioned at the $k$-th index across all layers is defined according to a stick-breaking process: 
\begin{align}
    \pi^{0}_{k} &= \prod^{k}_{i=1} v_{i}, \quad v_{k} \sim \textrm{Beta}(\alpha, 1), \quad k = 1, \cdots \infty.
\end{align}
Child IBPs are defined over the structure of each individual hidden layer of a BNN which depend on $\pi^{0}_{k}$ to define the respective Bernoulli probabilities of selecting neuron $k$ in layer $j$:
\begin{align}
    \pi_{jk} &\sim \textrm{Beta}\left(\alpha_j \pi^{0}_{k}, \alpha_{j}(1- \pi_{k}^{0})\right), \label{eq:child_beta}
\end{align}
for $j \in \{1, \cdots J$\},  $k \in \{1, \cdots \infty \}$, where $J$ is the number of layers in a BNN, $\alpha_j$ are hyperparameters \citep{thibaux2007hierarchical,gupta2012slice}. The selection of the $k$-th neuron in the $j$-th layer by a particular data point $i$ in the dataset of size $N$ is thus:
\begin{align}
    z_{ijk} &\sim \textrm{Bern}(\pi_{jk}), \, i = 1, \cdots N. \label{eq:hibp_bern}
\end{align}
Notice that if $k$ is small, $\pi^{0}_{k}$ is close to $1$ then the shape parameter of the child Beta distribution will be large. At the same time the scale parameter will be small. So the Bernoulli probability in Equation~(\ref{eq:hibp_bern}) will be close to $1$, as $k$ increases $\pi^{0}_{k}$ and $\pi_{jk}$ decrease. To infer the posterior $p(\bm{v}, Z, \bm{w}|\mathcal{D})$, we perform SSVI.

\footnotetext{The mean-field approximation removes all edges from these graphical models.}

\subsection{Structured VI}
\label{sec:hibp_svi}
A structured variational posterior distribution which retains properties of the true posterior is desired such that the global stick-breaking probabilities influence child stick-breaking probabilities of each layer of the BNN. Let us define the variational distributions for our hidden variables as follows,  $q(v^0_k)= \textrm{Beta}(\alpha^{0}_{k}, \beta^0_{k})$ and $q(\pi_{jk}) = \textrm{Beta}(\alpha_j \pi^{0}_{k}, \alpha_j (1- \pi_{k}^{0}))$, $\pi^0_k = \prod_{i=1}^{k}v^0_{i}$,  $q(z_{ijk}) = \textrm{Bern}(\pi_{jk})$ and the weights of the BNN are drawn from $q(\bm{w}_{jk}) = \mathcal{N}(\bm{\mu}_{jk}, \bm{\sigma}^2_{jk} \id)$.

The structured variational distribution is defined as follows
\begin{equation}
q(\phi) = \prod^{K}_{k=1}q(v^{0}_{k})\prod^{J}_{j=1}q(\pi_{jk}| v^{0}_{k})q(\bm{w}_{jk})\prod^{N}_{i=1}q(z_{ijk}|\pi_{jk})
\end{equation}
where $\phi=\{ \alpha^0_k, \beta^0_k, \bm{\mu}_{jk}, \bm{\sigma}_{jk}\}$ for all $j$ and $k$, up to the variational truncation, $K$. Having defined the structured variational distribution, the negative ELBO is:
\begin{equation}
\begin{aligned}
\mathcal{L}(\bm{\phi}, \mathcal{D}) &= \text{KL}(q(\bm{v}^0) || p(\bm{v}^0)) + \sum^{J}_{j=1} \text{KL}(q(\bm{\pi}_j | \bm{v}^0) || p(\bm{\pi}_j | \bm{v}^0)) \\
& + \sum^{J}_{j=1}\text{KL}(q(\bm{w}_j) || p(\bm{w}_j)) \\
    & - \sum^{J}_{j=1}\sum_{i=1}^{N} \E_{q(\mathbf{\phi})}[\log p(y_i | \mathbf{x}_i, \mathbf{z}_{ij}, \bm{w}_j)] \label{eq:objective_func}\\
     & + \sum^{J}_{j=1}\sum_{i=1}^{N}\text{KL}(q(\mathbf{z}_{ij} | \bm{\pi}_j) || p(\mathbf{z}_{ij} | \bm{\pi}_j)).
\end{aligned}
\end{equation}
The child stick-breaking variational parameters for each layer are conditioned on the global stick-breaking parameters and the binary masks $z_{ijk}$ for each neuron $k$ in each layer $j$ are conditioned on the child stick-breaking variational parameters. Thus, the variational structured posterior is able to capture dependencies of the prior. The learnable parameters are $\alpha^0_{k}$, $\beta^0_{k}$, $\bm{\mu}_{jk}$ and $\bm{\sigma}_{jk}$ for all $k$ neurons and for all layers $j$.

\looseness=-1

\subsubsection{Inference}

The variational posterior is obtained by optimising Equation~(\ref{eq:objective_func}) using structured stochastic VI. For inference to be tractable, we utilise three reparameterisations. The first is for the Gaussian weights \citep{Kingma}. The second is an implicit reparameterisation of the Beta distribution \citep{Figurnov}. The third reparameterisation uses a Concrete relaxation to the Bernoulli distribution \citep{Maddison, Jang}. Details of these are in the Supplementary material, Section \ref{sec:model_inference}.

\looseness=-1

\section{Related Works}

\label{sec:related_works}
\paragraph{IBP priors and model selection in deep learning.} An IBP prior has been used in VAEs to automatically learn the number of latent features. Stick-breaking probabilities have been placed directly as the VAE latent state \citep{Nalisnick}. The IBP prior has been used to learn the number of features in a VAE hidden state using mean-field VI \citep{Chatzis} with black-box VI \citep{Ranganath2014} and structured VI \citep{Hoffman2015,Singh}. As an alternative to truncation, \cite{Xu2019} use a Russian roulette sampling scheme to sample from the infinite sum in the stick-breaking process for the IBP. Model selection for BNNs has been performed with the Horse-shoe prior over weights \citep{Ghosh2019}. The IBP prior has been employed in BNNs to induce sparcity \citep{Panousis2019} and simultaneously to our work for CL \citep{Kumar}. Both of these approaches apply the IBP prior differently to our work and previous work applying the IBP to VAEs. Also \cite{Kumar} deviates from sequential Bayes by storing masks over weights for each task and uses design choices which mean the IBP is not the sole means of selecting weights, we expand on this in Section~\ref{sec:comparison}. Instead of using an IBP prior, Bernoulli distributions (or its Concrete relaxation) is used to select the width and number of layers in a BNN \citep{Dikov2019}. This approach is not non-parametric and so not as desirable for CL. Recently and subsequently to our work, \cite{Mehta2020} have proposed a CL approach where weights matrices are factorised and the IBP prior applied to the diagonal in the factorisation. This work does not use sequential Bayes for CL but rather different neural network weights are used for different tasks and thus alleviates forgetting.

\paragraph{Bayesian continual learning.} Repeated application of Bayes' rule can be used to update a model given the arrival of a new task. Previous work has used Laplace approximations \citep{Kirkpatrick} and variational inference \citep{VCL}. Bayesian methods can also be intuitively thought of as a weight space regularisation. Explicit regularisation in weight space have also proved successful in CL \citep{Zenke2017, Schwarz2018}. Our method builds upon \cite{VCL} as the framework for learning continually. None of these works deal with the issue of resource allocation to alleviate potential overfitting or underfitting problems in CL. The model we present adapts its size for CL, the scenario where BNNs need adapt to a changing data distribution or to \emph{concept drift} in CL has been studied too \citep{Kurle2020}. Bayesian CL can be interpreted as a regularisation in weight space. However, intuitively it is more sensible to regularise NN function outputs between tasks \citep{Benjamin2019, titsias2019functional, Rudner2021cfsvi}.


\paragraph{Adaptive models in continual learning.} Non-Bayesian CL approaches use additional neural resources to learn new tasks and remember previous tasks. One approach boils down to learning individual networks for each task \citep{Rusu}. More efficient use of resources can be done by selective retraining of neurons and expansion with a group sparsity regulariser \citep{Yoon}. However this approach is unable to shrink and continues to expand if it overfits on the first task. Another approach uses reinforcement learning by adding neural resources by penalising the complexity of the task network in the reward function \citep{Xu2018}. Recently \cite{Rao2019} propose an unsupervised CL model (CURL) in scenarios that lack a task identifier. CURL is adaptive, insofar that if a new task is detected then a new component is added to the mixture of Gaussians in the model. 
Task learning in CL can be modeled as a mixture of expert models. The experts are distributed according to a Dirichlet prior \citep{Lee2020}; new experts can be added to the mixture automatically. 

\section{Experimental Results}
\label{sec:results}

\begin{figure}
    \centering
    \includegraphics[width=0.49\textwidth]{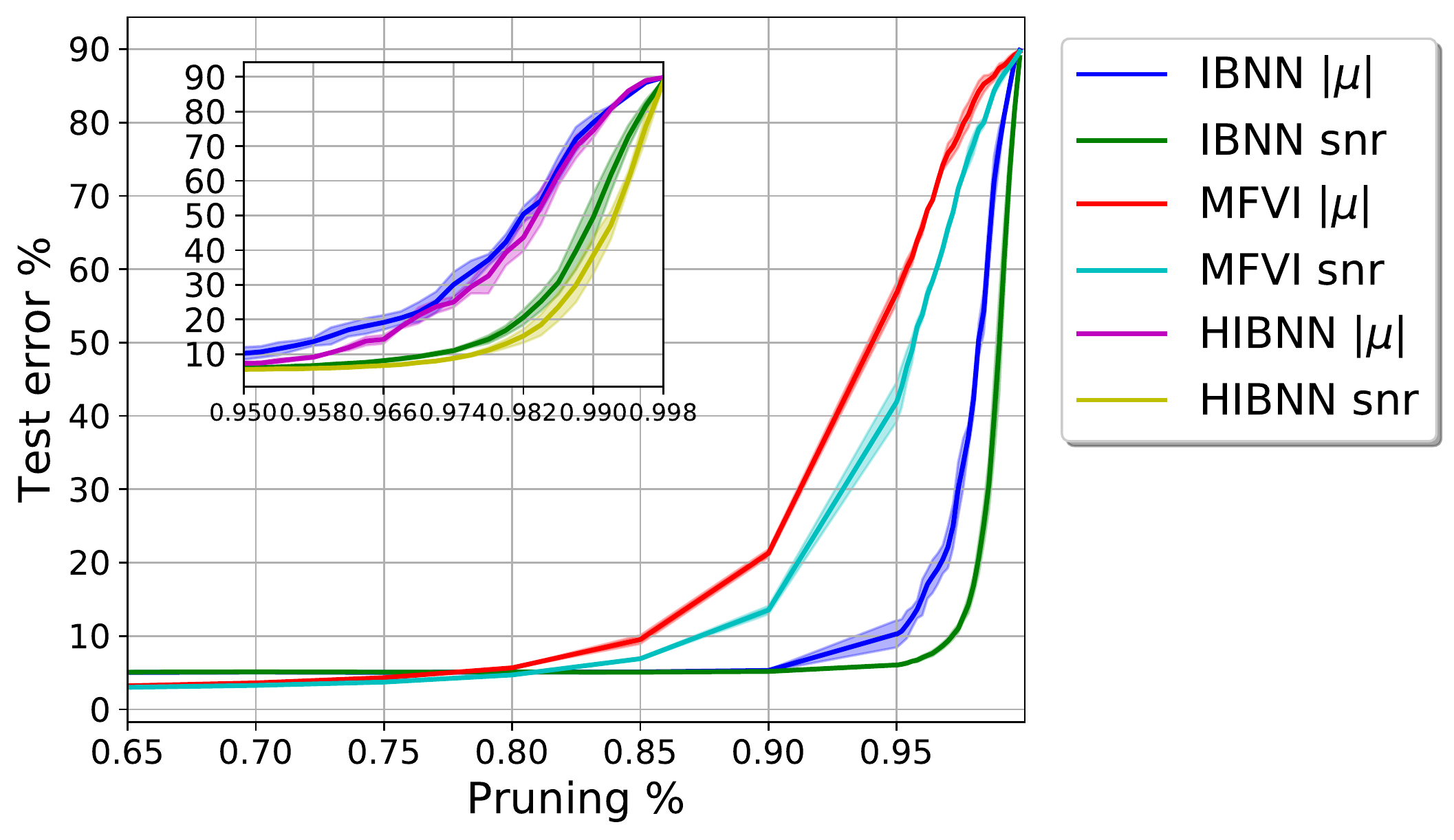}
    \caption{The weight pruning curves show test error versus the percentage of weights which have been zeroed out according to the magnitude of the variational mean and snr ($|\mu| / \sigma$). The HIBNN and IBNN are much sparser than a MFVI BNN and the HIBNN is more robust to pruning and therefore sparser than the IBNN.}
    \label{fig:wp}
\end{figure}

To demonstrate the effectiveness of the IBP and H-IBP priors on determining the size of the BNN, we perform weight pruning to see whether the pruned weights coincide with the weights dropped by the IBP and H-IBP priors in Section~\ref{sec:wp}. Furthermore, we then use the IBNN and HIBNN in a CL setting in Section~\ref{sec:results_cl}. Also in the supplementary material, further continual learning results and all experimental details are outlined. Unless explicitly stated, all curves are an average of 5 independent runs $\pm$ one standard error.  By test error, we refer to $(1 - \text{accuracy}) \times 100$.\footnote{Our code is available at \url{https://github.com/skezle/IBP_BNN}.}

\subsection{IBP induces sparsity}
\label{sec:wp}
\begin{figure}
     \centering
     \begin{subfigure}[t]{0.23\textwidth}
         \centering
         \includegraphics[width=0.99\textwidth]{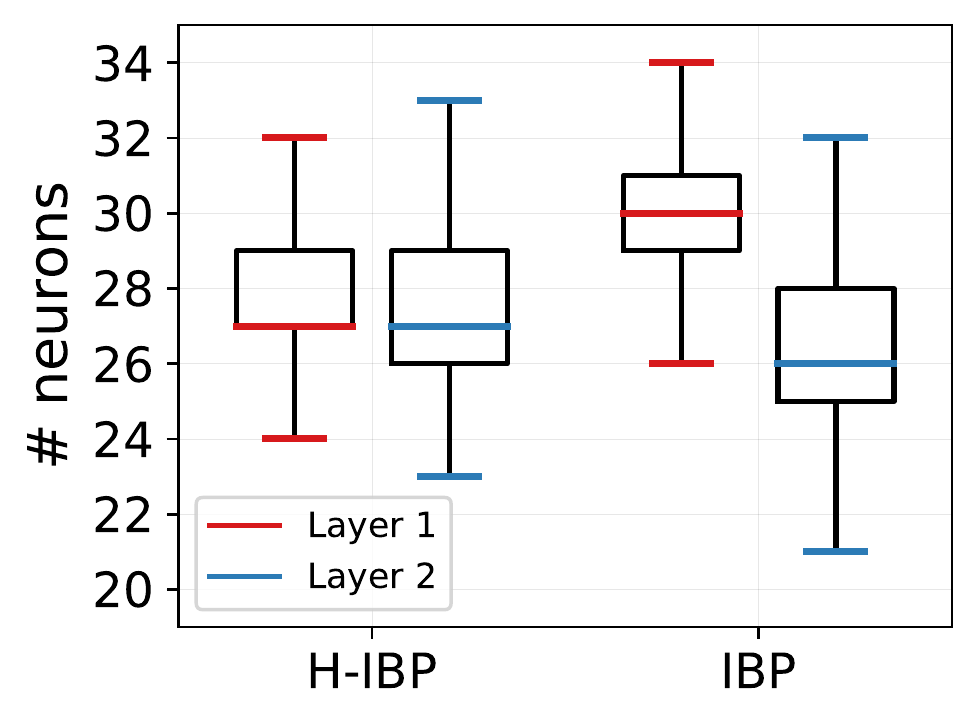}
          \label{fig:hibp_vs_ibp_wp_bp}
     \end{subfigure}
     \hfill
     \begin{subfigure}[t]{0.23\textwidth}
         \centering
         \includegraphics[width=0.94\textwidth]{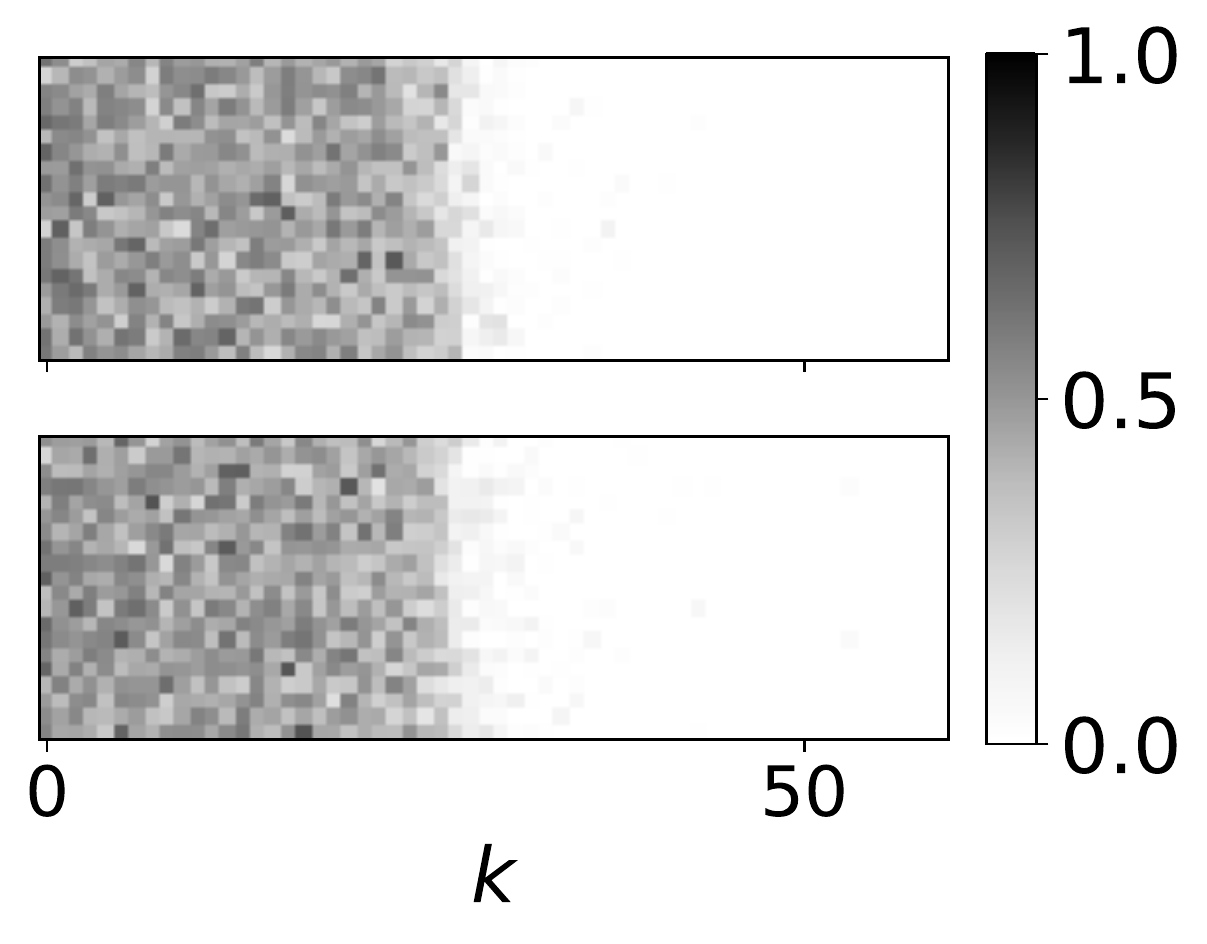}
          \label{fig:hibp_mnist_Zs}
     \end{subfigure}
        \caption{The number of active neurons in each layer of the IBNN and HIBNN. The HIBNN introduces an inductive bias which encourages and enables a regular structure. The $Z$ matrices for the HIBNN show the neurons which are being learnt on the left.\protect\footnotemark}
        \label{fig:wp_artefacts}
\end{figure}

We perform weight pruning to see whether the IBP posterior sensibly selects neurons through the binary matrix $Z$. Weights are pruned in two ways. The first is pruning according to $|\mu|$: zeroing out weights according to the magnitude of their mean. Important weights will be large in absolute value and so pruned last. Secondly, according to signal to noise ratio: $|\mu| / \sigma$ (snr). Weights with high uncertainty will also be zeroed out first. Weight pruning is performed on MNIST and compared to a mean-field BNN \citep{Blundell} (denoted MFVI in plots). The pruning accuracies in Figure~\ref{fig:wp} demonstrate that the HIBNN is indeed much sparser than a BNN and that pruning according to snr is more robust, as expected. The HIBNN is more robust to pruning than the IBNN due to its inductive bias leading to the more regular structure Figure~\ref{fig:wp_artefacts}. The baseline BNN has two layers with hidden state sizes of $200$, the HIBNN and IBNN use a variational truncation of $K=200$ for fair comparison. The HIBNN and IBNN achieves an accuracy of $0.95$ before pruning while MFVI achieves $0.98$ before pruning. This gap in performance is due to the approximate inference of the H-IBP and IBP posteriors and the various reparameterisations used, in particular the Concrete reparameterisation which is applying a `soft' mask on the hidden layers of the HIBNN. The IBNN and HIBNN are slightly less sparse compared to Sparse Variational Dropout which is specifically designed to be sparse \citep{Molchanov2017}, see Section~\ref{sec:pruning_svd} in the supplement.

\begin{figure}
     \centering
     \includegraphics[width=0.49\textwidth]{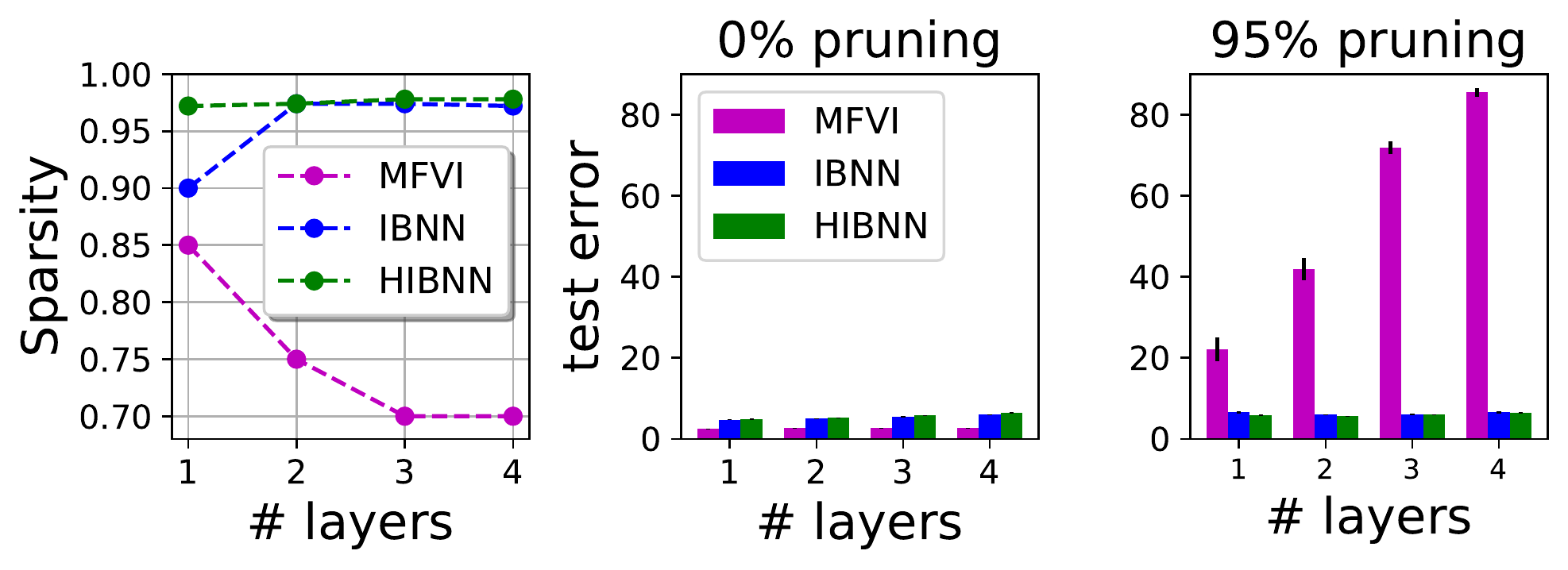}
     \caption{\textbf{Left}, as the depth of the IBNN and HIBNN increases the networks tend to remain very sparse while the MFVI BNN is becomes less sparse. \textbf{Middle \& Right}, the HIBNN and IBNN remain robust to pruning with increasing depth.}
      \label{fig:mnist_depth_sparsity}
\end{figure}

\textbf{Varying depth.} As we increase the depth we see that the HIBNN and IBNN remain sparse while the MFVI's sparsity decreases, Figure~\ref{fig:mnist_depth_sparsity}. We measure the sparsity as the pruning percentage at which the accuracy drops over $10\%$ (the kinks in Figure~\ref{fig:mnist_depth_sparsity}). There is little variation of the accuracy with depth for all models before pruning, however after pruning $95\%$ of the weights with the snr our models retain their performance while the MFVI BNN performs worse with depth since it becomes less sparse with depth Figure~\ref{fig:mnist_depth_sparsity}. See the supplement for the same analysis on fashion-MNIST, Section~\ref{sec:depth_fmnist_app}.
\looseness=-1
\looseness=-1
\footnotetext{$Z$ is usually shown in \textit{left-ordered form}, however since the inference procedure is based on the stick breaking construction, order is meaningful and sorted according to $k$ \citep{Xu2019}.}

\subsection{Continual learning experiments}
\label{sec:results_cl}

\begin{figure}
     \centering
     \includegraphics[width=0.40\textwidth]{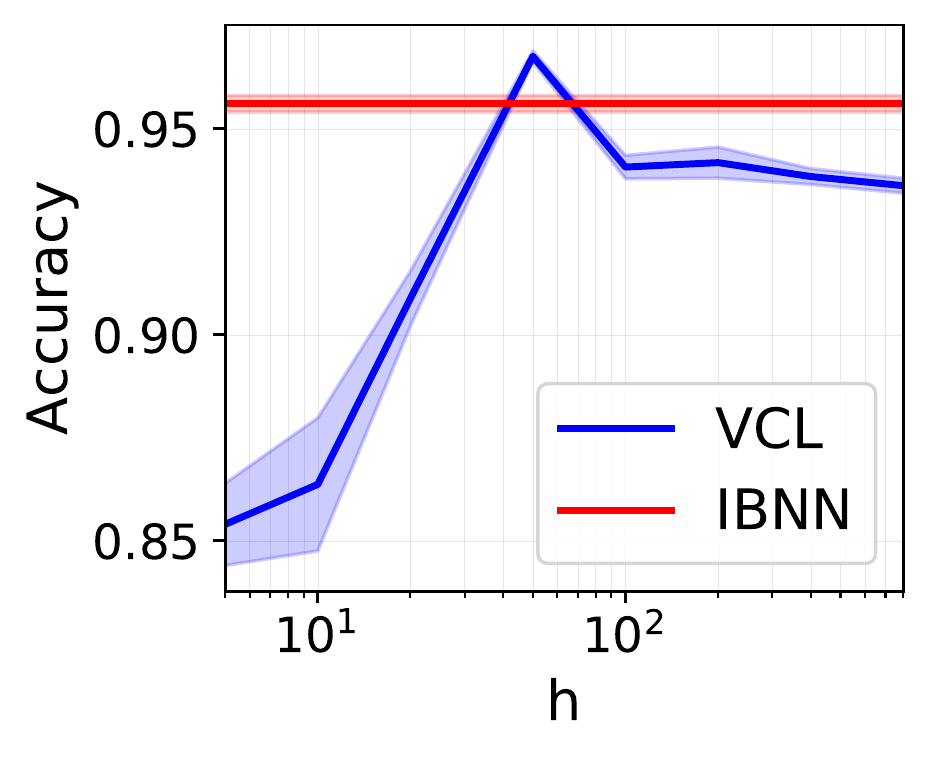}
     \caption{Test accuracies for Permuted MNIST \texttt{CL1}, for different VCL widths. Our model yields good results without having to specify a width, only a small range of values outperforms our model.}
      \label{fig:accs_perm_h_study}
\end{figure}

\begin{figure}
     \centering
     \includegraphics[width=0.42\textwidth]{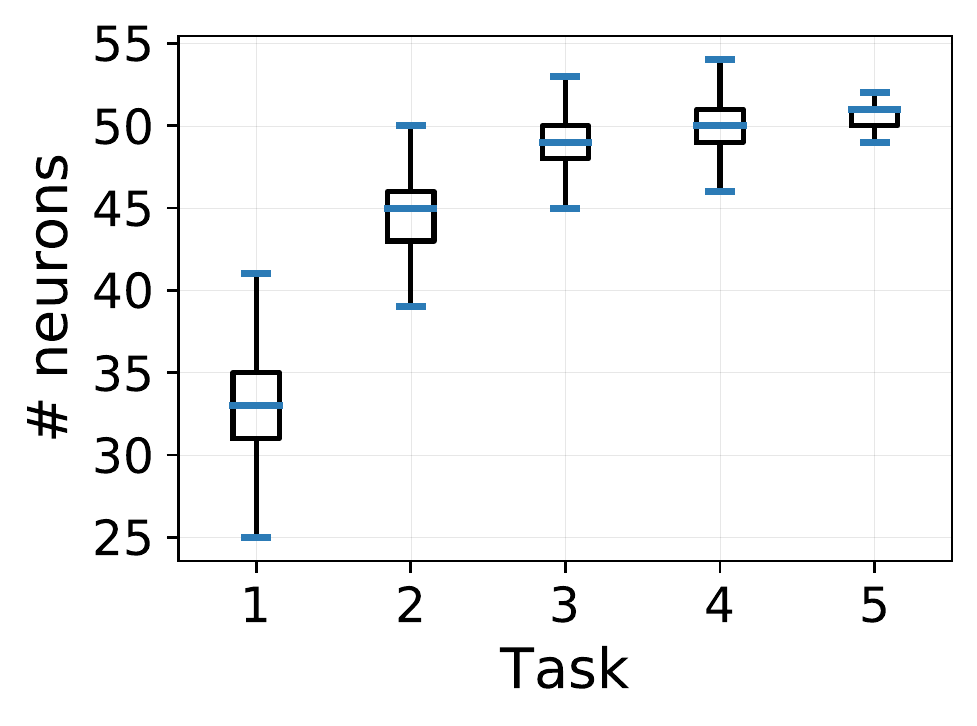}
     \caption{Number of active neurons\protect\footnotemark   selected by the IBP variational posterior for each task, for permuted MNIST \texttt{CL1}. Our model is able adapt and manage resources effectively.}
      \label{fig:Zs_bp_cl1_perm}
\end{figure}

\paragraph{Adaptive complexity.} Approximate inference of the IBP and H-IBP posteriors is challenging in a stationary setting making the performance attenuated in comparison to a BNN with only independent Gaussian weights. Despite this, the approximate IBP and H-IBP posteriors are useful in non-stationary CL setting, where the amount of resources are unknown beforehand. In Figure~\ref{fig:accs_perm_h_study}, one can see that the average accuracies across all CL tasks for permuted MNIST vary considerably with the hidden state size for VCL hence the benefit of our model which automatically infers the hidden state size for each task, see Figure~\ref{fig:Zs_bp_cl1_perm}. The details of the experiment will be introduced below.
\looseness=-1
\looseness=-1
\paragraph{Continual learning scenarios.} Three different CL scenarios are used for evaluation \citep{VanDeVen}. The first is \textit{task incremental learning} ($\texttt{CL1}$) where the task identifier is given during evaluation. The second is \textit{domain incremental learning} (\texttt{CL2}), the task identifier needs to be inferred at test time. The domain increases with each new task and the models are required to perform binary classification. The third is \textit{incremental class learning} (\texttt{CL3}), the task identifier and specific class need to be inferred at test time. The models are required to do multi-class classification for each task. During training, the task identifiers are given for all scenarios.\footnote{\emph{Task-free CL} is a more challenging scenario and requires the model the infer new tasks during training.} Using a multi-head architecture and the predictive entropy we can infer the task; the head with the lowest predictive entropy is chosen for \texttt{CL2} and \texttt{CL3} \citep{VonOswald2020}. For Permuted MNIST \texttt{CL2} a single-head architecture is used \citep{VanDeVen}.

\paragraph{Baseline models.} We compare our models to VCL \citep{VCL}, since the IBNN and HIBNN models build on top of it. We also compare to EWC \citep{Kirkpatrick} and SI \citep{Zenke2017}. We compare to DEN \citep{Yoon} which is an expansion method which expands a neural network by a fixed number of weights for each new task, uses regularisation to mitigate overfitting and freezes weights from previous tasks. For DEN the predictive entropy is used for inferring the correct head for \texttt{CL2} and \texttt{CL3}. Another baseline used is GEM \citep{Lopez-Paz}, which uses replay as its primary mechanism for alleviating forgetting. We hypothesise that GEM will be insensitive to model size. It also achieves strong results \citep{Hsu2018}. VCL is the fairest comparison for our model and will also utilise uncertainties for \texttt{CL2} and \texttt{CL3}. Our objective is to demonstrate that limited or excessive neural resources can cause problems in CL in comparison to our adaptive model.
\looseness=-1
\looseness=-1
\paragraph{MNIST Experimental details.} All models use a single layer with varying hidden state sizes. The use of a single layer is enough as MNIST is a simple task. The results for EWC in Table~\ref{tab:mnist_accs} on Split MNIST outperform those presented in \cite{Hsu2018, VanDeVen} which use larger models. We report accuracies for our non adaptive baselines (EWC, SI, GEM and VCL) over a set of hidden state sizes $\mathcal{H} = \{10, 50, 100, 400\}$. A hidden state of $10$ might seem small but we also set the IBP prior parameter $\alpha=5$ for task $1$. This corresponds to only $5$ neurons being selected by each data point in expectation. The initial hidden state size for DEN is set to $50$.

\paragraph{Permuted MNIST benchmarks.} The permuted MNIST benchmark involves performing multiclass classification on MNIST where in each task, the pixels have been shuffled by a fixed permutation. Our model is able to overcome overfitting and underfitting which result in increased forgetting which affect VCL. See Figure~\ref{fig:accs_all_perm_mh} in the supplement, for a per task accuracy breakdown. In contrast, the IBNN expands continuously Figure~\ref{fig:Zs_bp_cl1_perm}. There is a small gap in performance between our model and VCL for $h=50$ due to the approximations used for inference of the variational IBP posterior. Our method outperforms all regularisation based methods and DEN\footnote{It is not clear how to reconcile single-head networks and DEN, thus \texttt{CL2} for DEN is omitted.} on all CL scenarios and provides comparable results to GEM, see Table~\ref{tab:mnist_accs}. 

\paragraph{Split MNIST benchmarks.} The split MNIST benchmark for CL involves a sequence of classification tasks of MNIST and more difficult variants with background noise and background images denoted S+$\epsilon$ and S+img. For EWC, SI and VCL notice a considerable difference in performance with hidden state size in Table~\ref{tab:mnist_accs}. GEM is sensitive for $\texttt{CL3}$ only. EWC and SI perform well for \texttt{CL1} only. The IBNN outperforms VCL as it is not susceptible to overfitting or underfitting and thus propagating a subsequent poor posterior for a new task resulting in forgetting.

Split MNIST is a simple task which doesn't show overfitting, hence the use of the MNIST variant datasets where the IBNN outperforms all VCL models of different sizes as it is not susceptible to overfitting or underfitting. Indeed our method outperforms not only VCL but EWC and SI and performs comparably to GEM. When analysing the performance of the VCL baselines, we notice they have a tendency to overfit on the second task and propagate a poor approximate posterior and hence underperform in comparison to the IBNN model, Figures~\ref{fig:accs_all_random_mh} and \ref{fig:accs_all_split_images_mh} in the supplement. The IBNN increases its capacity over the course of CL, Figure \ref{fig:Zs_bp_cl1_normal_random_mnist} in the supplement. The standard errors for VCL and our method on $\texttt{CL3}$ are large due to the severity of making mistakes for multiclass classification.
\looseness=-1
\looseness=-1
\footnotetext{We define a neuron as active by aggregating all neurons where $z_{ik} > 0.1$ for data point $\bf{x}_i$ and neuron $k$.}

\begin{table*}
  \centering
  \caption{Average test accuracies on MNIST and variants over 5 runs. For EWC, SI, GEM and VCL the median accuracy is taken from hidden state sizes, $\mathcal{H} = \{10, 50, 100, 400 \}$. We also show the range between max and min average accuracies in $\mathcal{H}$. The models with the best median or mean accuracy are highlighted. If the IBNN mean accuracy lies within the min-max range then our model is also highlighted. Our IBNN achieves good performance overall compared to the baselines which can underfit/overfit. DEN performs very poorly on permuted MNIST.}

\begin{tabular}{lllll|llll}
\toprule
  &  EWC \scalebox{0.85}{(\texttt{max}, \texttt{min})} & SI \scalebox{0.85}{(\texttt{max}, \texttt{min})} & GEM \scalebox{0.85}{(\texttt{max}, \texttt{min})} & DEN & VCL \scalebox{0.85}{(\texttt{max}, \texttt{min})} & IBNN \\
  \midrule
 P \texttt{CL1} & $90.0$ \scalebox{0.85}{$(94.0, 79.1)$} & $91.8$ \scalebox{0.85}{$(95.1, 82.9)$} & $94.8$ \scalebox{0.85}{$(95.9, 88.4)$} & $91.4$\scalebox{0.85}{$\pm 0.5$} & $93.9$ \scalebox{0.85}{$(96.8, 86.9)$} & $\bf{95.6}$\scalebox{0.85}{$\pm \bf{0.2}$} \\
 P \texttt{CL2} & $88.2$ \scalebox{0.85}{$(93.0, 78.2)$} & $89.2$ \scalebox{0.85}{$(93.3, 84.1)$} & $\bf{95.6}$ \scalebox{0.85}{$\bf{(96.7, 88.1)}$} & - & $88.7$ \scalebox{0.85}{$(95.1, 75.2)$} & $\bf{93.7}$\scalebox{0.85}{$\pm \bf{0.6}$} \\
 P \texttt{CL3} & $59.3$ \scalebox{0.85}{$(66.9, 24.0)$} & $47.5$ \scalebox{0.85}{$(55.4, 18.1)$} & $\bf{94.6}$ \scalebox{0.85}{$\bf{(95.8, 87.3)}$} & $63.9$\scalebox{0.85}{$\pm 19.2$} & $84.0$ \scalebox{0.85}{$(94.1, 40.5)$} & $\bf{93.8}$\scalebox{0.85}{$\pm \bf{0.3}$} \\

 \midrule
 S \texttt{CL1}  & $98.9$ \scalebox{0.85}{$(99.0, 94.8)$} & $98.1$ \scalebox{0.85}{$(99.2, 95.5)$} & $98.1$ \scalebox{0.85}{$(98.2, 98.0)$} & $\bf{99.1}$\scalebox{0.85}{$\pm \bf{0.1}$}  & $96.6$ \scalebox{0.85}{$(98.0, 93.7)$} & $95.3$\scalebox{0.85}{$\pm 2.0$} \\
 S \texttt{CL2}  & $63.7$ \scalebox{0.85}{$(74.8, 63.3)$} & $78.0$ \scalebox{0.85}{$(80.1, 72.9)$} & $94.0$ \scalebox{0.85}{$(94.8, 92.2)$} & $\bf{98.9}$\scalebox{0.85}{$\pm \bf{0.1}$} & $87.4$ \scalebox{0.85}{$(94.9, 81.9)$} & $91.0$\scalebox{0.85}{$\pm 2.2$} \\
 S \texttt{CL3}  & $19.9$ \scalebox{0.85}{$(21.8, 19.8)$} & $18.8$ \scalebox{0.85}{$(19.6, 15.4)$} & $89.2$ \scalebox{0.85}{$(89.6, 87.8)$} & $\bf{99.1}$\scalebox{0.85}{$\pm \bf{0.1}$} & $69.0$ \scalebox{0.85}{$(78.9, 66.3)$} & $85.5$\scalebox{0.85}{$\pm 3.2$} \\
 \midrule

 S+$\epsilon$ \texttt{CL1}  & $94.6$ \scalebox{0.85}{$(96.5, 81.8)$} & $88.4$ \scalebox{0.85}{$(91.1, 72.7)$} & $95.6$ \scalebox{0.85}{$(95.7, 95.1)$} & $\bf{97.2}$\scalebox{0.85}{$\pm \bf{0.2}$} & $89.2$ \scalebox{0.85}{$(90.3, 86.3)$} & $95.1$\scalebox{0.85}{$\pm 1.1$} \\
 
 S+$\epsilon$ \texttt{CL2}  & $72.7$ \scalebox{0.85}{$(74.6, 68.7)$} & $66.1$ \scalebox{0.85}{$(72.1, 61.8)$} & $78.2$ \scalebox{0.85}{$(78.7, 77.2)$} & $84.8$\scalebox{0.85}{$\pm 16.7$} & $69.1$ \scalebox{0.85}{$(70.2, 61.7)$} & $\bf{89.7}$\scalebox{0.85}{$\pm \bf{3.8}$} \\
 
 S+$\epsilon$ \texttt{CL3}  & $18.8$ \scalebox{0.85}{$(19.0, 13.7)$} & $13.7$ \scalebox{0.85}{$(16.7, 10.4)$} & $74.1$ \scalebox{0.85}{$(74.4, 70.7)$} & $\bf{90.9}$\scalebox{0.85}{$\pm \bf{12.8}$} & $32.5$ \scalebox{0.85}{$(37.8, 30.0)$} & $78.7$\scalebox{0.85}{$\pm 11.7$} \\
 \midrule
 
 S+img \texttt{CL1}  & $88.2$ \scalebox{0.85}{$(90.0, 78.1)$} & $80.8$ \scalebox{0.85}{$(87.4, 67.8)$} & $91.7$ \scalebox{0.85}{$(91.8, 91.1)$} & $\bf{93.8}$\scalebox{0.85}{$\pm \bf{0.8}$}  & $87.1$ \scalebox{0.85}{$(87.9, 85.7)$} & $91.6$\scalebox{0.85}{$\pm 1.2$} \\
 S+img \texttt{CL2}  & $67.7$ \scalebox{0.85}{$(73.9, 62.9)$} & $63.3$ \scalebox{0.85}{$(67.0, 58.5)$} & $\bf{74.1}$ \scalebox{0.85}{$\bf{(76.4, 74.0)}$} & $\bf{79.1}$\scalebox{0.85}{$\pm \bf{13.1}$}  & $70.4$ \scalebox{0.85}{$(75.2, 65.3)$} & $\bf{80.5}$\scalebox{0.85}{$\pm \bf{7.8}$} \\
 S+img \texttt{CL3}  & $17.0$ \scalebox{0.85}{$(18.0, 12.0)$} & $13.4$ \scalebox{0.85}{$(16.4, 10.3)$} & $63.9$ \scalebox{0.85}{$(64.5, 56.3)$} & $\bf{91.6} $\scalebox{0.85}{$\pm \bf{5.1}$}  & $54.3$ \scalebox{0.85}{$(66.1, 39.9)$} & $66.2$\scalebox{0.85}{$\pm 13.4$} \\
 \bottomrule
\end{tabular}
\label{tab:mnist_accs}
\end{table*}

DEN performs well on all Split MNIST tasks and variants due to its ``time-stamped inference'' which segregates parts of the network per task and so uncertainties over seen tasks are well defined thus the good results for \texttt{CL2} and \texttt{CL3}. Indeed removing it renders performance comparable to IBNN, see Section~\ref{sec:den_no_ts} in the supplement. Statistical processes which mimic this could be an interesting direction for Bayesian expansion methods in CL.

\begin{table}
\centering
\caption{Accuracies for tasks of increasing difficulty. Accuracies are an average over $5$ runs. VCL use different hidden state sizes, $\mathcal{H} = \{50, 100, 400\}$, we show the range between the max and min average accuracies centre at the median. Our models are able to overcome underfitting.}
\begin{tabular}{llll}
    \toprule
      & VCL \scalebox{0.99}{(\texttt{max}, \texttt{min})} & IBNN & HIBNN\\
      \midrule
     \texttt{CL1} & $\bf{82.3}$ \scalebox{0.99}{$\bf{(86.1, 81.2)}$} & $80.8$\scalebox{0.99}{$\pm 2.0$} & $\bf{81.3}$\scalebox{0.99}{$\pm \bf{1.8}$} \\
     \texttt{CL2} & $79.9$ \scalebox{0.99}{$(83.1, 76.7)$} & $78.7$\scalebox{0.99}{$\pm 1.5$} & $\bf{81.5}$\scalebox{0.99}{$\pm \bf{1.4}$}\\
     \bottomrule
    \end{tabular}
     \label{tab:exp_accs}
     \vspace{-3.0mm}
\end{table}

\begin{figure}
     \centering
     \includegraphics[width=0.5\textwidth]{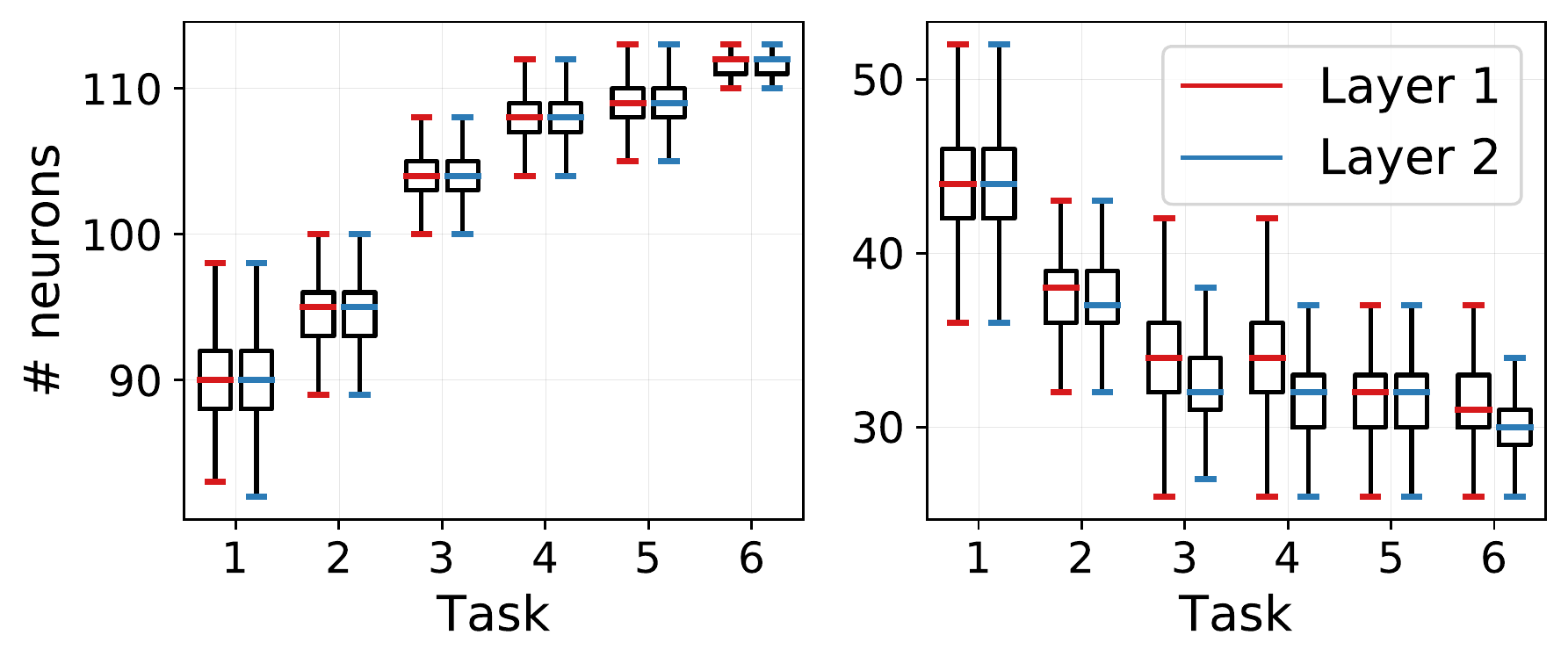}
    \caption{Inferred structure for the HIBNN and IBNN respectively. Our models can infer varied structures, expand and contract.}
    \label{fig:bp_mix}
\end{figure}

\paragraph{Increasing task complexity.} To test the expansion capabilities of our models we devise a set tasks of increasing difficulty: two tasks from MNIST, followed by two from fashion MNIST, followed by two from CIFAR10. We compare the HIBNN and IBNN models to VCL with two layers and widths in $\mathcal{H} = \{50, 100, 400\}$. Our models have two layers with $K=200$. The larger width VCL networks perform well but smaller ones exhibit forgetting due to underfitting. For the HIBNN we allow the hyperparameter $\alpha_j$ to increase for each new dataset seen i.e. every two tasks. We perform random search over the IBP and H-IBP parameters for the IBNN and HIBNN models, Section~\ref{sec:rs}. The HIBNN performs better than the IBNN, additionally we can see Figure~\ref{fig:bp_mix} that both of our models can have very different structure after learning on a task. Notice that since there is a sharing of parameters at a global level that the widths of the HIBNN match across different layers unlike the IBNN.

\section{Conclusion and Discussion}
\label{sec:discussion}

Model size is an important contributing factor for CL performance. Most CL methods assume a perfectly selected model. Our novel Bayesian CL framework nonparametrically adapts the complexity of a BNN to the task difficulty. Our model is based on the IBP prior for selecting the number of neurons for each task and uses stochastic variational inference. The models presented reconcile two different approaches to CL: Bayesian or regularization based approaches and dynamic architecture approaches through the use of a IBP and H-IBP prior. 

We have demonstrated our model on MNIST like datasets. Future work will focus on showing that our method is able to scale to larger vision datasets which are used in CL. This will involve applying the IBP and H-IBP priors to Convolutional Neural Networks (CNN). Also ensuring that inference can be performed efficiently will be another challenge since inference in the IBNN and HIBNN is more expensive than for VCL (see Section~\ref{sec:training_times}) and VCL has been shown to produce poor results when scaling to these larger models \citep{Pan2020}. This could be performed with other inference methods such as using natural gradients \citep{Osawa2019} or only training certain parts of the convolutional block of the CNN \citep{Ovadia2019}. This is left for future work.

\begin{acknowledgements} The authors would like to than the anonymous reviewers for the helpful comments. The authors would also like to thank Sebastian Farquhar for helpful discussions and Kyriakos Polymenakos for proof reading a draft of the paper. SK is funded by an Oxford-Man studentship. Thanks for to all the great open source software which made this research possible \texttt{Numpy} and \texttt{scipy} \citep{2020SciPy-NMeth}, \texttt{matplotlib} \citep{Hunter:2007} \texttt{Tensorflow} \citep{Abadi:2016:TSL:3026877.3026899} and \texttt{Tensorflow Distributions} \citep{Dillon2017}.
\end{acknowledgements}


\bibliography{kessler_287.bbl}

\newpage
\onecolumn
\appendix

\section*{Appendices}

We present details of inference for our model in Section~\ref{sec:model_inference} and additional experiments.

\section{Model and Inference Details}
\label{sec:model_inference}
In this section, we present the variational BNN with the structure of the hidden layer determined by the H-IBP variational posterior (a.k.a. HIBNN). It is straightforward to apply the following methodology for a simpler model with independent IBP variational posteriors determining the structure of each hidden layer (referred to the IBNN in the main paper).

We derive a structured variational posterior where dependencies are established between global parameters and local parameters \citep{Hoffman2015}. Once the variational posterior is obtained, we can follow the VCL framework \citep{VCL} for CL. The following set of equations govern the stick-breaking H-IBP model for an arbitrary layer $j \in \{1,...,J\}$ with $k$ neurons in each layer of a BNN:
\begin{align}
    v_k &\sim \textrm{Beta} \left(\alpha, 1 \right), &&\textrm{for} \, k = 1, \ldots ,\infty, \\
    \pi^0_k & = \prod^{k}_{i=1}v_i, &&\forall \, k, \\
    \pi_{jk} &\sim \textrm{Beta}(\alpha_j \pi_{k}^{0}, \alpha_j (1-\pi^{0}_{k})), &&\forall \, k, j = 1, \ldots, J, \\
    z_{ijk} & \sim \textrm{Bern}(\pi_{jk}), && \forall \, j, k, i = 1, \ldots, N, \label{eq:ibp_bern} \\
    \bm{w}_{jk} & \sim \mathcal{N}(\bm{\mu}_{jk}, \bm{\sigma}_{jk}^2 \id), && \forall \, j, k,\\
    h_{jk} &= f(\bm{h}_{j-1} \bm{w}_{jk}) \circ z_{ijk} && \forall \, i, j, k.
\end{align}
The index $k$ denotes a particular neuron, $j$ denotes a particular layer, and $\bm{w}_{jk}$ denotes a column (of $W_j$) from the weight matrix mapping hidden states from layer $j-1$ to layer $j$, such that $h_j = f(h_{j-1} W_j) \circ Z_j$. We denote $\circ$ as the elementwise multiplication operation. The binary matrix $Z_j$ controls the inclusion of a particular neurons in layer $j$. The dimensionality of our variables are as follows $\bm{w}_{jk} \in \R^{k_{j-1}}$, $h_{j-1} \in \R^{k_{j-1}}$, $h_{j} \in \R^{k_{j}}$ and $z_{ijk} \in \Z_{2} = \{0,1 \}$.

The closed-form solution to the true posterior of the H-IBP parameters and BNN weights involves integrating over the joint distribution of the data and hidden variables, $\phi = \{ \bm{Z}, \bm{\pi}, \bm{\pi^0}, \bm{w} \}$. Since it is not possible to obtain a closed-form solution to this integral, variational inference together with reparameterisations of the variational distributions are used \citep{Kingma,Figurnov} to employ gradient based methods. The variational approximation used is 
\begin{align}
    q(\phi) &= \prod^{K}_{k=1} q(v^0_k; \alpha^0_{k}, \beta^0_{k})\prod^{J}_{j=1}q(\pi_{jk}|v_{k}^{0}) q(\bm{w}_{jk}; \bm{\mu}_{jk}, \bm{\sigma}_{jk})\prod_{i=1}^{N}q(z_{ijk}|v^0_{k}),\label{eq:variational_q}
\end{align}
where the variational posterior is truncated up to $K$, the prior is still infinite \citep{Nalisnick}. The set of variational parameters which we optimise over are $\bm{\varphi} = \{\bm{\alpha^0}, \bm{\beta^0}, \bm{\mu}, \bm{\sigma}\}$ and $q(\phi)$ is the variational distribution. Each term in Equation~(\ref{eq:variational_q}) is specified as follows
\begin{align}
\label{eq:ibp}
    q(v^0_k; \alpha^0_{k}, \beta^0_{k}) &= \text{Beta}(v_k; \alpha^0_{k}, \beta^0_{k}), \\
    \pi^0_k &= \prod^{k}_{i=1} v^0_i, \\
    q(\pi_{jk}| v^0_k) &= \text{Beta}(\pi_{jk}; \alpha_j \pi_{k}^{0}, \alpha_j (1-\pi^{0}_{k}))\\
    q(z_{ijk}| v^0_k) &= \text{Bern}(z_{ijk}; \pi_{jk}), \label{eq:ibp_bern_v} \\
    q(\bm{w}_{jk}; \bm{\mu}_{jk}, \bm{\sigma}_{jk}) &= \mathcal{N}( \bm{w}_{jk}; \bm{\mu}_{jk}, \bm{\sigma}^2_{jk} \id) \label{eq:ibp_gauss_v},
\end{align}
where $\alpha_j$ are hyperparameters \citep{gupta2012slice}. Now that we have defined our structured variational approximation in Equation~(\ref{eq:variational_q}) we can write down the objective 
\begin{align}
    \quad \argmin \KL \left( q(\bm{\phi}) || p(\bm{\phi} | \mathcal{D}) \right)
    = \argmin \KL \left( q(\bm{\phi}) || p(\bm{\phi}) \right) - \E_{q(\bm{\phi})}[\log p(\mathcal{D} | \bm{\phi})].
\end{align}
In the above formula, $q(\bm{\phi})$ is the approximate posterior and $p(\bm{\phi})$ is the prior. By substituting Equation~(\ref{eq:variational_q}), we obtain the negative ELBO objective:
\begin{align}
\begin{split}
\label{eq:elbo}
    \mathcal{L}(\bm{\phi}, \mathcal{D}) &= \textrm{KL}(q(\bm{v^0})|| p(\bm{v^0})) 
    + \sum^{J}_{j=1}\textrm{KL}(q(\bm{\pi}_j | \bm{v}^0) || p(\bm{\pi}_j | \bm{v}^0)) + \textrm{KL}(q(\bm{w_j})|| p(\bm{w_j})) \\
    &- \sum^{J}_{j=1}\sum_{i=1}^{N} \E_{q(\bm{\phi})}[\log p(y_i | \bm{x_i}, \bm{z_{ij}}, \bm{w_j})] 
    + \sum^{J}_{j=1}\sum_{i =1}^{N} \textrm{KL}(q(\bm{z_{ij}} | \bm{\pi_j})||p(\bm{z_{ij}} | \bm{\pi_j})).
\end{split}
\end{align}

Estimating the gradient of the Bernoulli and Beta variational parameters requires a suitable reparameterisation. Samples from the Bernoulli distribution in Equation~(\ref{eq:ibp_bern_v}) arise after taking an \texttt{argmax} over the Bernoulli parameters. The \texttt{argmax} is discontinuous and a gradient is undefined. The Bernoulli is reparameterised as a Concrete distribution \citep{Maddison, Jang}. Additionally the Beta is reparameterised implicitly \citep{Figurnov} to separate sampling nodes and parameter nodes in the computation graph and allow the use of stochastic gradient methods to learn the variational parameters $\bm{\varphi}$ of the approximate H-IBP posterior. The mean-field approximation for the Gaussian weights of the BNN is used, $\bm{w}$ in Equation~(\ref{eq:ibp_gauss_v}) \citep{Blundell, VCL}. In the next sections we detail the reparameterisations used to optimise Equation~(\ref{eq:elbo}).

\subsection{The variational Gaussian weight distribution reparameterisation}
\label{sec:gauss_reparam}
The variational posterior over the weights of the BNN are diagonal Gaussian $\bm{w}_{jk} \sim \mathcal{N}(\bm{w}_{jk} | \bm{\mu}_{jk}, \bm{\sigma}^2_{jk} \id)$. By using a reparameterisation, one can represent the BNN weights using a deterministic function  $\bm{w}_{jk} = g(\epsilon; \bm{\varphi})$, where $\epsilon~\sim \mathcal{N}(0, \id)$ is an auxiliary variable and $g(\epsilon; \bm{\varphi})$ a deterministic function parameterised by $\bm{\varphi} = (\bm{\mu}_{jk}, \bm{\sigma}_{jk})$, such that:
\begin{align}
\label{eq:reparam_gauss}
    \nabla_{\varphi} \E_{q(\bm{w}_{jk}; \varphi)}\left[ f(\bm{w}_{jk}) \right] &= \E_{q(\epsilon)}\left[ \nabla_{\varphi}f(\bm{w}_{jk}) \right]|_{\bm{w}_{jk} = g(\epsilon; \varphi)},
\end{align}
where $f$ is an objective function, for instance Equation~(\ref{eq:elbo}).
The BNN weights can be sampled directly through the reparameterisation: $\bm{w}_{jk} = \bm{\mu}_{jk} + \bm{\sigma}_{jk} \epsilon$. By using this simple reparameterisation the weight samples are now deterministic functions of the variational parameters $\bm{\mu}_{jk}$ and $\bm{\sigma}_{jk}$ and the noise comes from the independent auxiliary variable $\epsilon$ \citep{Kingma}. Taking a gradient of our ELBO objective in Equation~(\ref{eq:elbo}) the expectation of the log-likelihood may be rewritten by integrating over $\epsilon$ so that the gradient with respect to $\bm{\mu}_{jk}$ and $\bm{\sigma}_{jk}$ can move into the expectation according to Equation~(\ref{eq:reparam_gauss}) (the backward pass). In the forward pass, $\epsilon \sim q(\epsilon)$ is sampled to compute $\bm{w}_{jk} = \bm{\mu}_{jk} + \bm{\sigma}_{jk} \epsilon$.

\subsection{The implicit Beta distribution reparameterisation} 
\label{sec:beta_reparam}
Implicit reparameterisation gradients \citep{Figurnov, Mohamed2019MonteCG} is used to learn the variational parameters for Beta distribution. \cite{Nalisnick} propose to use a Kumaraswamy reparameterisation. However, in CL a Beta distribution  rather than an approximation is desirable for repeated Bayesian updates.

There is no simple inverse of the reparameterisation for the Beta distribution like the Gaussian distribution presented earlier. Hence, the idea of implicit reparameterisation gradients is to differentiate a standardisation function rather than have to perform its inverse. The standardisation function is given by the Beta distribution's CDF. 

Following \cite{Mohamed2019MonteCG}, the derivative required for general stochastic VI is:
\begin{align}
\begin{split}
    \nabla_{\varphi} \E_{q(v^0; \varphi)}\left[ f(v^0) \right] &= \E_{q(\epsilon)}\left[ \nabla_{\varphi}f(v^0) \right]|_{v^0 = g(\epsilon; \varphi)} \\
    &= \E_{q(v^0; \varphi)}[\nabla_{v^0}f(v^0)\nabla_{\varphi}v^0],
\end{split}
\end{align}


where $f$ is an objective function, i.e. Equation~(\ref{eq:elbo}), $\varphi = \{\alpha^0_k , \beta^0_k\}$ are the Beta parameters. The implicit reparameterisation gradient solves for $\nabla_{\varphi}v^0$, above, using implicit differentiation \citep{Figurnov}:
\begin{align}
\label{eq:implicit_grad}
    \nabla_\varphi v^0  = - ( \nabla_z g^{-1}_\varphi(v^0;\varphi))^{-1} \nabla_\varphi g^{-1}_\varphi(v^0;\varphi),
\end{align}

where $v^0 = g(\epsilon; \varphi)$, $g(.)$ is the inverse CDF of the Beta distribution and $\epsilon\sim\textrm{Unif}[0,1]$ and so $\epsilon = g^{-1}(v^0;\varphi)$ is the standardisation path which removes the dependence of the distribution parameters on the sample. The key idea in implicit reparameterisation is that this expression for the gradient in Equation~(\ref{eq:implicit_grad}) only requires differentiating the standardisation function $g^{-1}_\varphi(v^0;\varphi)$ and not inverting it. Given $v^0 = g(\epsilon; \varphi) = F^{-1}(v^0;\varphi)$ then using Equation~(\ref{eq:implicit_grad}) the implicit gradient is:
\begin{align}
\begin{split}
    \label{eq:implicit_grad_univariate}
    \nabla_\varphi v^0  &= \nabla_\varphi F^{-1}(v^0;\varphi) \\
    &= \frac{\nabla_\varphi F(v^0 ; \varphi)}{p_\varphi (v^0;\varphi)},
\end{split}
\end{align}
where $p_\varphi (v^0;\varphi)$ is the Beta PDF. The CDF of the Beta distribution, is given by 
\begin{align}
    F(v^0;\varphi )=\frac{B(v^0;\varphi)}{B(\varphi)}
\end{align}
where $B(v^0;\varphi)$ and $B(\varphi)$ are the incomplete Beta function and Beta function, respectively.  The derivatives of $\nabla_\varphi F(v^0 ; \varphi)$ do not admit simple analytic expressions. Thus, numerical approximations have to be made, for instance, by using Taylor series for $B(v^0;\varphi)$ \citep{jankowiak2018pathwise,Figurnov}.

In the forward pass, $v^0 \sim q_\varphi(v^0) $ is sampled from a Beta distribution or alternatively from two Gamma distributions. That is, if $v^0_1 \sim \textrm{Gamma}(\alpha^0_k, 1)$ and $v^0_2 \sim \textrm{Gamma}(\beta^0_k, 1)$, then $\frac{v^0_1}{v^0_1+v^0_2} \sim \textrm{Beta}(\alpha^0_k, \beta^0_k)$.

Then, in the backward pass we estimate the partial derivatives w.r.t. $\varphi=\{ \alpha^0_k, \beta^0_k \}$ as can be taken by using Equation~(\ref{eq:implicit_grad_univariate}) and previous equations. Implicit reparameterisation of the Beta distribution is readily implemented in \texttt{TensorFlow Distributions} \citep{Dillon2017}.

\subsection{The variational Bernoulli distribution reparameterisation}
\label{sec:concrete_reparam}

The Bernoulli distribution can be reparameterised using a continuous relaxation to the discrete distribution and so Equation~(\ref{eq:reparam_gauss}) can be used. 

Consider a discrete distribution $(\alpha_1, \cdots \alpha_K)$ where $\alpha_j \in \{0, \infty\}$ and $D \sim \text{Discrete}(\alpha) \in \{0,1\}$, then $P(D_j = 1) = \frac{\alpha_j}{\sum_k \alpha_k}$. Sampling from this distribution requires performing an $\texttt{argmax}$ operation, the crux of the problem is that the $\texttt{argmax}$ operation doesn't have a well defined derivative.

To address the derivative issue above, the Concrete distribution \citep{Maddison} or Gumbel-Softmax distribution \citep{Jang} is used as an approximation to the Bernoulli distribution. The idea is that instead of returning a state on the vertex of the probability simplex like $\texttt{argmax}$ does, these relaxations return states inside the inside the probability simplex (see Figure 2 in \cite{Maddison}). The Concrete formulation and notation from \cite{Maddison} are used. We sample from the probability simplex using
\begin{align}
\label{eq:X_concrete}
X_j = \frac{\exp((\log \alpha_j + G_j)/\lambda)}{\sum^{n}_{i=1}\exp((\log \alpha_i + G_i)/\lambda)},
\end{align}
with temperature hyperparameter $\lambda \in (0, \infty)$, parameters $\alpha_j \in (0, \infty)$ and i.i.d. Gumbel noise $G_j \sim \text{Gumbel}(0, 1)$. This equation resembles a \texttt{softmax} with a Gumbel perturbation. As $\lambda \rightarrow 0$ the \texttt{softmax} computation approaches the $\texttt{argmax}$ computation. This can be used as a relaxation of the variational Bernoulli distribution and can be used to reparameterise Bernoulli random variables to allow gradient based learning of the variational Beta parameters downstream in our model. 

When performing variational inference using the Concrete reparameterisation for the posterior, a Concrete reparameterisation of the Bernoulli prior is required to properly lower bound the ELBO in Equation~(\ref{eq:elbo}). If $q(z_{ijk}; \pi_{jk} | v^0_k)$ is the variational Bernoulli posterior over sparse binary masks $z_{ijk}$ for weights $\bm{w}_{jk}$ and $p(z_{ijk}; \pi_{jk} | v^0_{k})$ is the Bernoulli prior. To guarantee a lower bound on the ELBO, both Bernoulli distributions require replacing with Concrete densities, i.e.,
\begin{align}
\label{eq:kl_concrete}
    \textrm{KL}\left[q(z_{ijk}; \pi_{jk} | v^0_k)||p(z_{ijk}; \pi_{jk} | v^0_k)\right] 
     \geq \textrm{KL}\left[q(z_{ijk}; \pi_{jk}, \lambda_1 | v^0_k)||p(z_{ijk}; \pi_{jk}, \lambda_2 | v^0_k)\right],
\end{align}
where $q(z_{ijk}; \pi_{jk}, \lambda_1 | v^0_k)$ is a Concrete density for the variational posterior with parameter $\pi_{jk}$, temperature parameter $\lambda_1$ given global parameters $v^0_k$. The Concrete prior is $p(z_{ijk}; \pi_{jk}, \lambda_2 | v^0_k)$. Equation~(\ref{eq:kl_concrete}) is evaluated numerically by sampling from the variational posterior (we will take a single Monte Carlo sample \citep{Kingma}).  At test time one can sample from a Bernoulli using the learnt variational parameters of the Concrete distribution \citep{Maddison}.

In practice, the log transformation is used to alleviate underflow problems when working with Concrete probabilities. One can instead work with $\exp(Y_{ijk}) \sim\text{Concrete}(\pi_{jk}, \lambda_1 | v^0_k)$, as the KL divergence is invariant under this invertible transformation and Equation~(\ref{eq:kl_concrete}) is valid for optimising our Concrete parameters \citep{Maddison}. For binary Concrete variables one can sample from $y_{ijk} = (\log \pi_{jk} + \log \epsilon - \log(1-\epsilon)) / \lambda_1$ where $\epsilon \sim \textrm{Unif}[0,1]$ and  the log-density (before applying the sigmoid activation) is $\log q(y_{ijk}; \pi_{jk}, \lambda_1 | v_{jk}) = \log \lambda_1 - \lambda_1 y_{ijk} + \log \pi_{jk} - 2\log(1+\exp(-\lambda_1 y_{ijk} + \log \pi_{jk}))$ \citep{Maddison}. The reparameterisation $\sigma(y_{ijk}) = g(\epsilon; \pi_{jk})$, where $\sigma$ is the sigmoid function, enables us to differentiate through the Concrete and use a similar formula to Equation~(\ref{eq:reparam_gauss}).

\section{Comparison of HIBNN and Sparse Variational Dropout}
\label{sec:pruning_svd}

\begin{figure}
     \centering
     \includegraphics[width=0.45\textwidth]{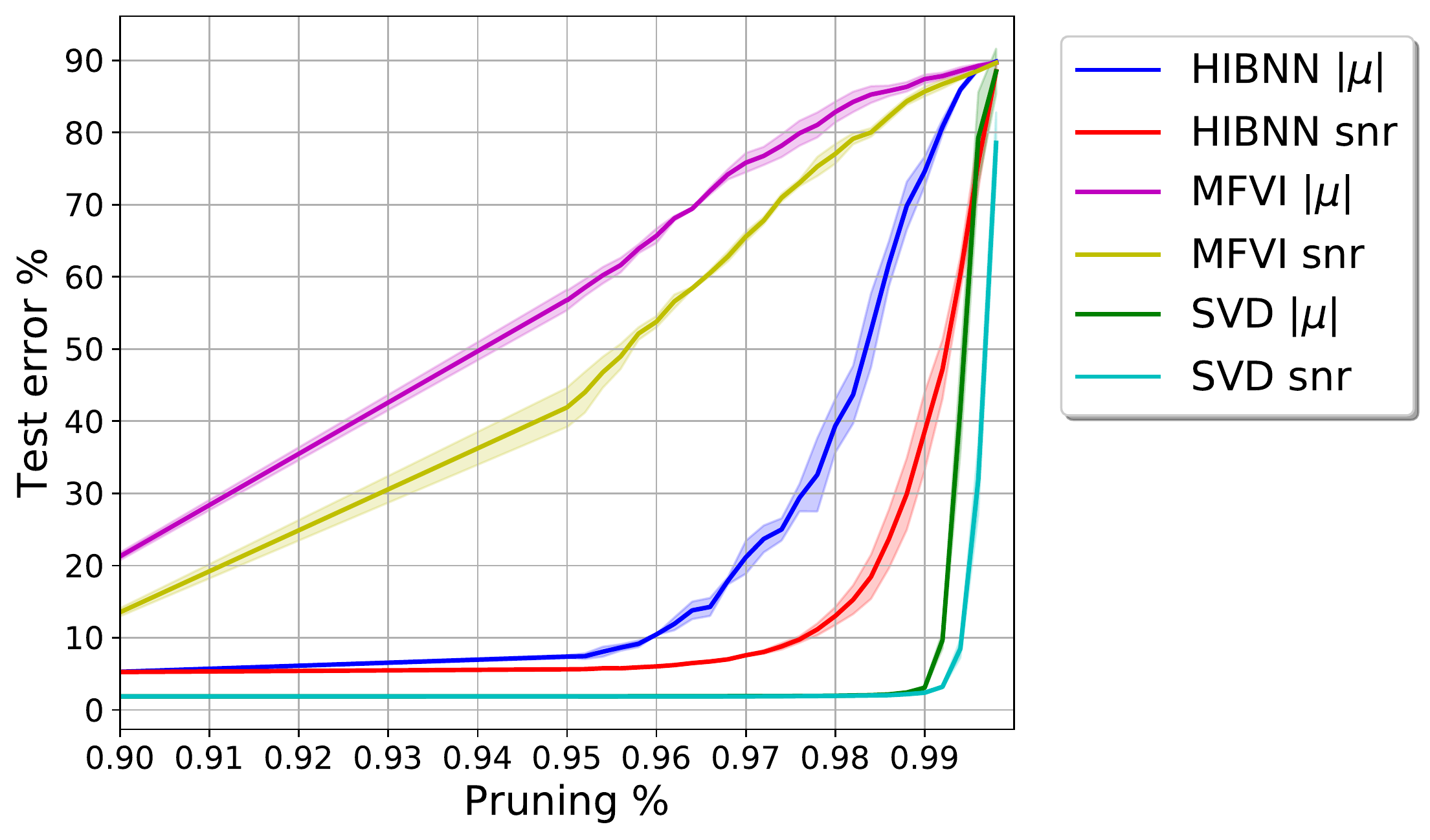}
     \caption{Test errors for different pruning cut-offs for the HIBNN and with a BNN trained with Sparse Variational Dropout (SVD) on MNIST. SVD is more sparse than the HIBNN.}
      \label{fig:wp_hibp_svd}
\end{figure}

Using the IBP and H-IBP priors for model selection induces sparsity as a side effect. Other priors such as Sparse Variational Dropout (SVD) \citep{Molchanov2017} or a horse-shoe prior on weights \citep{Louizos2017} are specifically employed for sparsity. By comparing the effect of weight pruning between a HIBNN and a BNN employing SVD, we can see that the HIBNN is not as sparse as SVD, although SVD is specifically designed to be sparse, unlike our method which employs a non-parametric prior for CL, but sparsity is a side effect. The accuracies obtained from SVD before pruning are $98.1$\scalebox{0.85}{$\pm 0.1$} compared to $95$\scalebox{0.85}{$\pm 0.0$}. The performance gap is due to the the difficulty in inference and with the variational approximations and reparameterisations, as noted when comparing to a BNN with no special prior in Section~\ref{sec:wp}. In terms of pruning, SVD is slightly more robust; performance starts to degrade after $99\%$ of weights are pruned in comparison to the HIBNN's $98\%$, see Figure~\ref{fig:wp_hibp_svd}.

The SVD BNN uses a two layer BNN with $200$ neurons and the HIBNN with a variational truncation of $K=200$ for a fair comparison. The H-IBP prior parameters and the initialisation of the variational posterior are  $\alpha^0_k = 4.2$ and $\beta_k^0 = 1.0$ for all $k$, the hyperparameter $\alpha_j = 4$ for all layers $j$. The Concrete temperatures used are $\lambda_1 = 0.7$ for the variational posterior and $\lambda_2 = 0.7$ for the Concrete prior. The prior on the Gaussian weights was set to $\mathcal{N}(0.0, 0.7)$ for these parameters where found to work well on split MNIST and were assumed to also work well for multiclass MNIST too.

Both networks are optimised using Adam \citep{Kingma2015} with a decaying learning rate schedule starting at $10^{-3}$ at a rate of $0.87$ every $1000$ steps, for $200$ epochs and using a batch size of $128$. Weight means are initialised with their ML parameters found after training for $100$ epochs and $\log \sigma^2 = -6$. Local reparameterisation \citep{LRP2015} is employed. SVD is trained for $100$ epochs while our method for $200$ epochs, as it requires more epochs to converge.
\section{Weight Pruning Further Results}

\subsection{MNIST}
\begin{figure}
     \centering
    \includegraphics[width=0.85\textwidth]{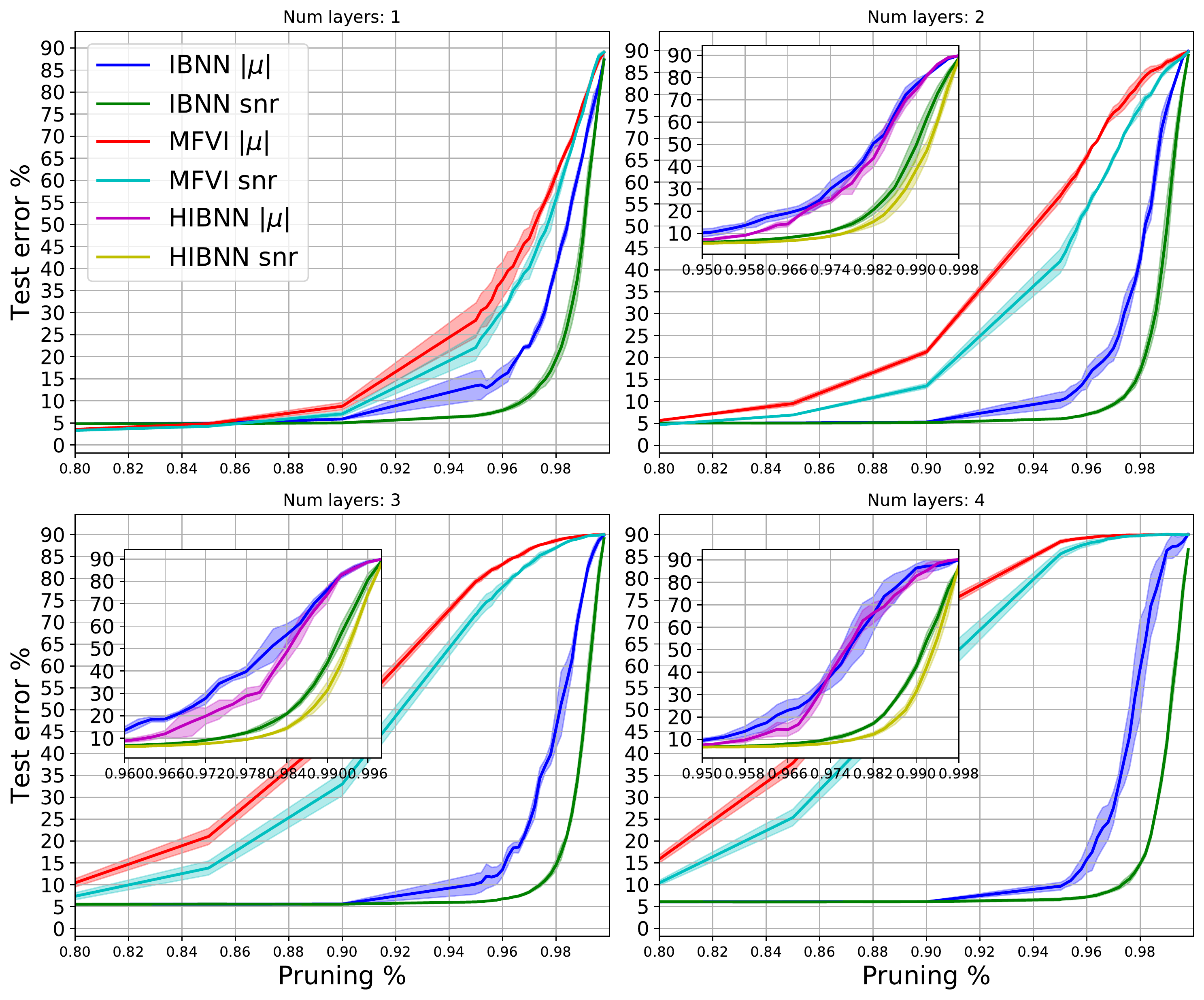}
     \caption{Weight pruning experiments for the IBNN and HIBNN on MNIST for models of different depth. As the depth increases the IBNN and HIBNN become more sparse.}
    \label{fig:accs_wp_mnist_all}
\end{figure}

To investigate the behaviour of the novel BNN introduced in this paper we perform simple and intuitive weight pruning experiments. The Gaussian weights of the BNN are zeroed out in two particular orders. The first is by the magnitude of the variational mean: $|\mu|$ the second is by the variational signal to noise ratio (snr): $|\mu| / \sigma$. For the IBNN and HIBNN, weights are pruned by according to these two metrics at the same time as having the sparse matrix $Z$ acting on each layer. As $Z$ is sparse by design we expect our models to be sparse in comparison to a BNN where the weights are modelled by mean-field variational inference (MFVI) \citep{Blundell}. Indeed the IBNN and HIBNN are sparse in comparison to MFVI and become sparser with increased depth. In Figure~\ref{fig:accs_wp_mnist_all} the weights are pruned and the accuracy on the test set is measured for networks of different sizes. MFVI is less sparse as the depth increases. Accuracy, does not vary for all models of depth $1$ to $4$. The MFVI models have a hidden state size of $200$ and the IBNN and HIBNN use a variational truncation of $K=200$ for fair comparison.
\label{sec:depth_mnist_app}

\subsection{Fashion-MNIST}
\label{sec:depth_fmnist_app}
We repeat this analysis for the fashion MNIST (fMNIST) \citep{xiao2017/online}. Similarly to MNIST the IBNN and HIBNN networks become sparser with increasing depth since the the drop off in accuracies arise for higher pruning percentages. There is little difference in the performance or sparsity between the IBNN and HIBNN.
\begin{figure}
     \centering
     \includegraphics[width=0.75\textwidth]{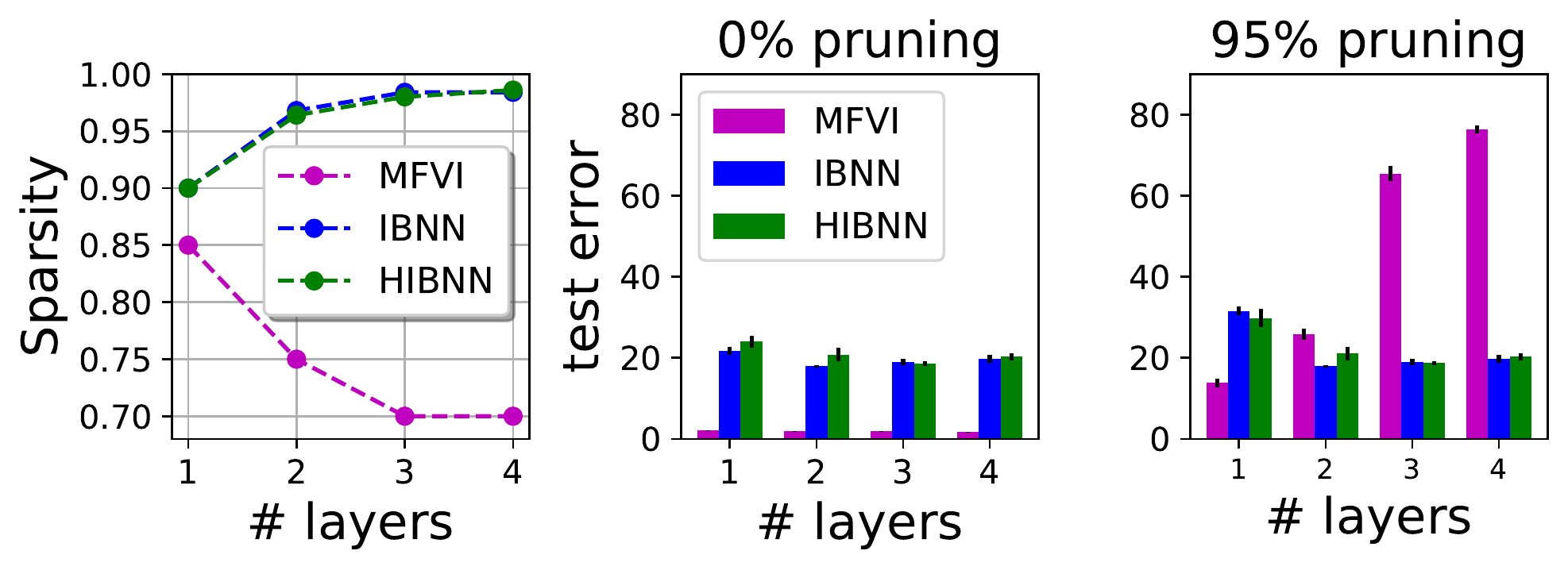}
     \caption{fMNIST network depth ablation. \textbf{Left}, as the depth of the IBNN and HIBNN increases the networks tend to remain very sparse while the MFVI BNN is becomes less sparse. \textbf{Middle \& Right}, the HIBNN and IBNN remain robust to pruning with increasing depth.}
      \label{fig:fmnist_depth_sparsity}
\end{figure}

\begin{figure}
     \centering
    \includegraphics[width=0.99\textwidth]{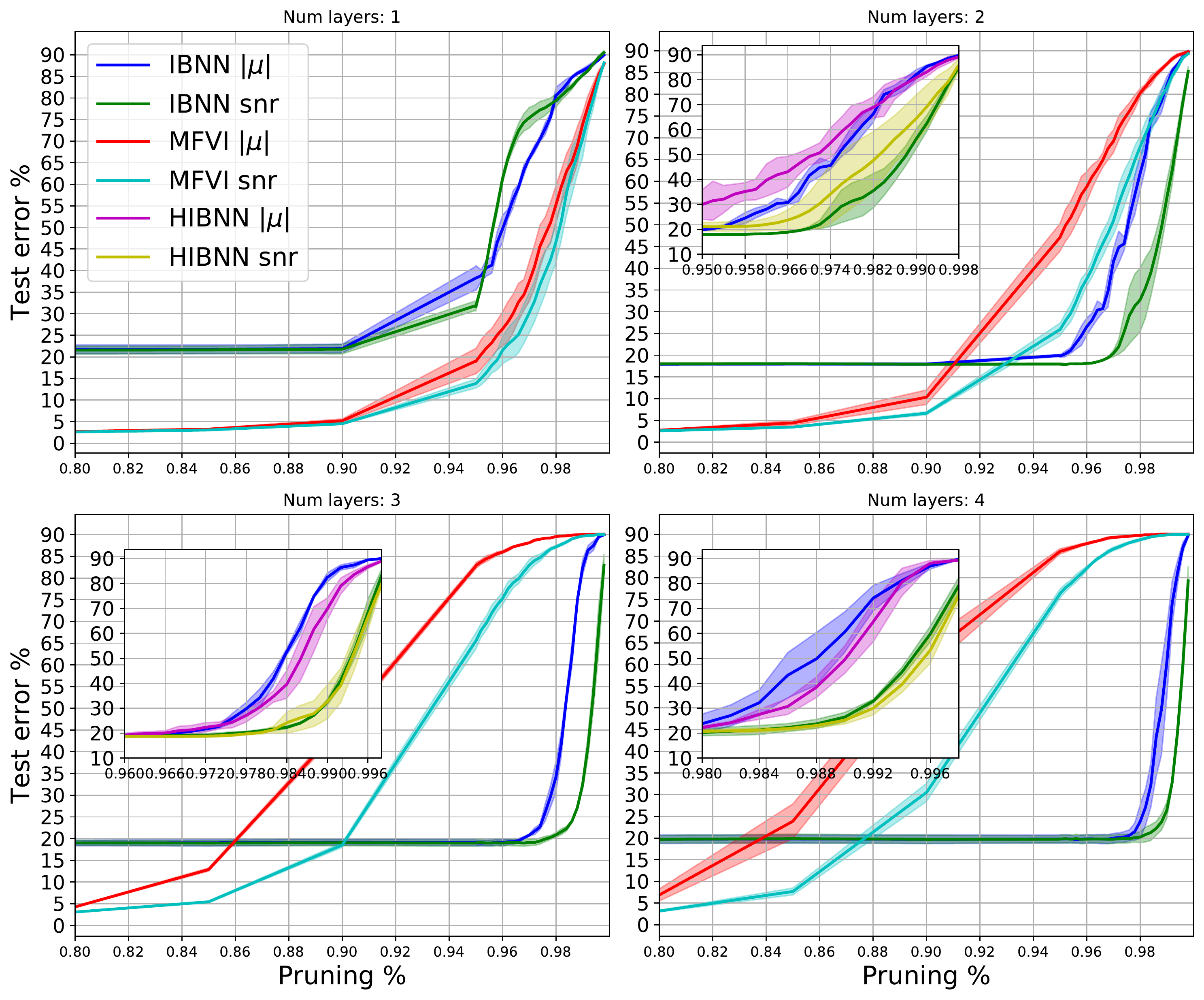}
     \caption{Weight pruning for the IBNN and HIBNN in comparison to MFVI on fashion-MNIST for models of different depth. As the depth increases the IBNN and HIBNN become more sparse.}
    \label{fig:accs_wp_fmnist_all}
\end{figure}

\section{Dynamically Expanding Networks and Time-Stamping}
\label{sec:den_no_ts}
DEN \citep{Yoon} ``time-stamps'' parts of its network allowing specific parts of the network to be used by a specific tasks thereby achieving good performance and good uncertainties over tasks which have been seen for \texttt{CL2} and \texttt{CL3}. This is at the cost of not allowing backward transfer from newly added task weights. Removing the time-stamping feature (and making DEN more inline with the IBNN) we see that it achieves similar performance for the more difficult MNIST variants, see Table~\ref{tab:mnist_accs_den}. This motivates segregating parts of a BNN to perform a specific task so as to mimic time-stamping for future research.

\begin{table}
  \centering
  \caption{Average test accuracy on MNIST CL experiments over 5 runs. After removing the time-stamping feature from DEN the IBNN achieves comparable performance. The best model between the IBNN and DEN with no time-stamping is highlighted.}
\begin{tabular}{ll|ll}
\toprule
  & DEN & DEN - no ts & IBNN \\
 \midrule
 P \texttt{CL1}  & $91.4$\scalebox{0.85}{$\pm 0.5$}  & $91.6$\scalebox{0.85}{$\pm 0.5$} & $\bf{95.6}$\scalebox{0.85}{$\pm \bf{0.2}$} \\
 P \texttt{CL3}  & $63.9$\scalebox{0.85}{$\pm 19.2$} & $75.1$\scalebox{0.85}{$\pm 21.0$} & $\bf{93.8}$\scalebox{0.85}{$\pm \bf{0.3}$} \\
 \midrule
 S \texttt{CL1}  & $99.1$\scalebox{0.85}{$\pm 0.1$}  & $\bf{98.8}$\scalebox{0.85}{$\bf{\pm 0.1}$} & $95.3$\scalebox{0.85}{$\pm 2.0$} \\
 S \texttt{CL2}  & $98.9$\scalebox{0.85}{$\pm 0.1$} & $\bf{98.9}$\scalebox{0.85}{$\pm \bf{0.1}$} & $91.0$\scalebox{0.85}{$\pm 2.2$} \\
 S \texttt{CL3}  & $99.1$\scalebox{0.85}{$\pm 0.1$} & $\bf{93.6}$\scalebox{0.85}{$\bf{\pm 10.4}$} & $\bf{85.5}$\scalebox{0.85}{$\bf{\pm 3.2}$} \\
 \midrule

 S+$\epsilon$ \texttt{CL1}  & $97.2$\scalebox{0.85}{$\pm 0.2$} & $\bf{95.1}$\scalebox{0.85}{$\bf{\pm 1.9}$} &$\bf{95.1}$\scalebox{0.85}{$\bf{\pm 1.1}$} \\
 
 S+$\epsilon$ \texttt{CL2}  & $84.8$\scalebox{0.85}{$\pm 16.7$} & $\bf{91.1}$\scalebox{0.85}{$\pm \bf{7.0}$} & $\bf{89.7}$\scalebox{0.85}{$\pm \bf{3.8}$} \\
 
 S+$\epsilon$ \texttt{CL3}  & $90.9$\scalebox{0.85}{$\pm 12.8$} & $54.1$\scalebox{0.85}{$\pm 24.1$} & $\bf{78.7}$\scalebox{0.85}{$\bf{\pm 11.7}$} \\
 \midrule
 S+img \texttt{CL1} & $93.8$\scalebox{0.85}{$\pm 0.8$}  & $\bf{91.7}$\scalebox{0.85}{$\bf{\pm 2.1}$} & $\bf{91.6}$\scalebox{0.85}{$\bf{\pm 1.2}$} \\
 S+img \texttt{CL2} & $79.1$\scalebox{0.85}{$\pm 13.1$}  & $\bf{98.9}$\scalebox{0.85}{$\pm \bf{0.1}$} & $80.5$\scalebox{0.85}{$\pm 7.8$} \\
 S+img \texttt{CL3} & $91.6$\scalebox{0.85}{$\pm 5.1$}  & $46.7$\scalebox{0.85}{$\pm 14.3$} & $\bf{66.2}$\scalebox{0.85}{$\pm \bf{13.4}$} \\
 \bottomrule
\end{tabular}
  \label{tab:mnist_accs_den}
\end{table}

\section{Overfitting and Underfitting in VCL}
\label{sec:app_overfit}

\paragraph{Preliminaries.} All task accuracies are an average of $5$ runs and by action neurons we mean that $z_{ijk} > 0.1$ for a datapoint $\bf{x}_i$ in layer $j$ and for neuron $k$.

\begin{figure*}
     \centering
     \begin{subfigure}{\textwidth}
        \includegraphics[width=1\textwidth]{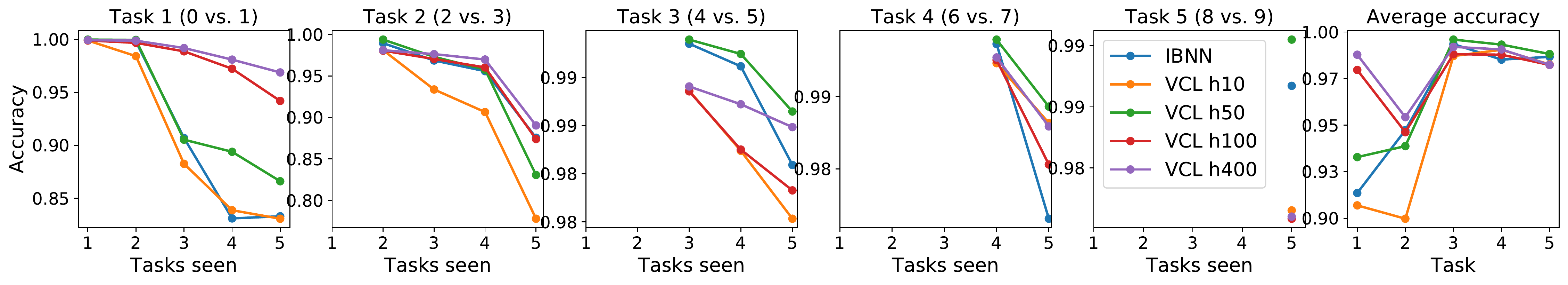}
     \end{subfigure}
     \caption{Split MNIST \texttt{CL1}, task accuracies versus the number of tasks the model has performed a Bayesian update. Our model is compared to VCL benchmarks with different numbers of hidden states denoted $hx, \, x \in \{10, 50, 100, 400\}$. For Split MNIST the tasks are simple and no overfitting is observed in VCL.}
      \label{fig:accs_all_normal_mh}
\end{figure*}

\begin{figure*}
     \centering
     \begin{subfigure}{\textwidth}
        \centering
        \includegraphics[width=1\textwidth]{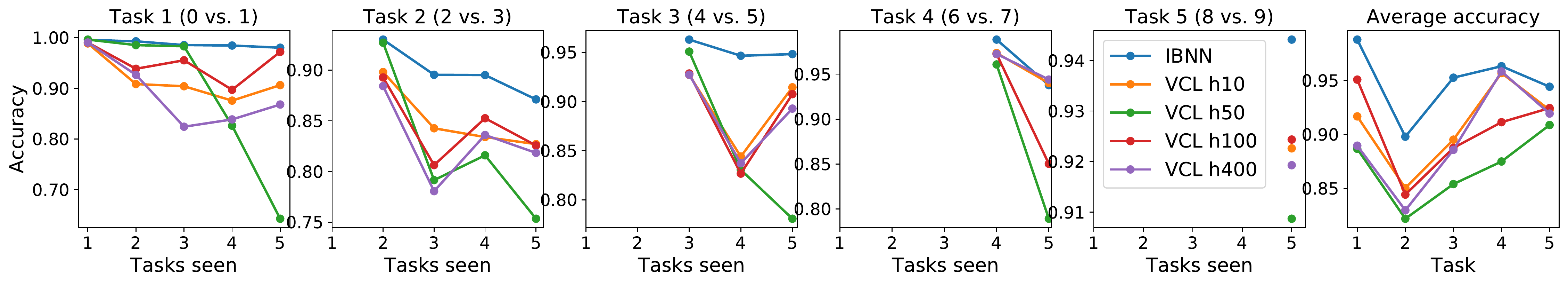}
     \end{subfigure}
     \caption{Split MNIST with random background noise for \texttt{CL1}, task accuracies versus the number of tasks the model has performed a Bayesian update. Our model is compared to VCL benchmarks with different numbers of hidden states denoted $hx, \, x \in \{10, 50, 100, 400\}$. Our model is able to counter overfitting issues and outperform VCL.}
      \label{fig:accs_all_random_mh}
\end{figure*}

\begin{figure}
     \centering
     \includegraphics[width=0.5\textwidth]{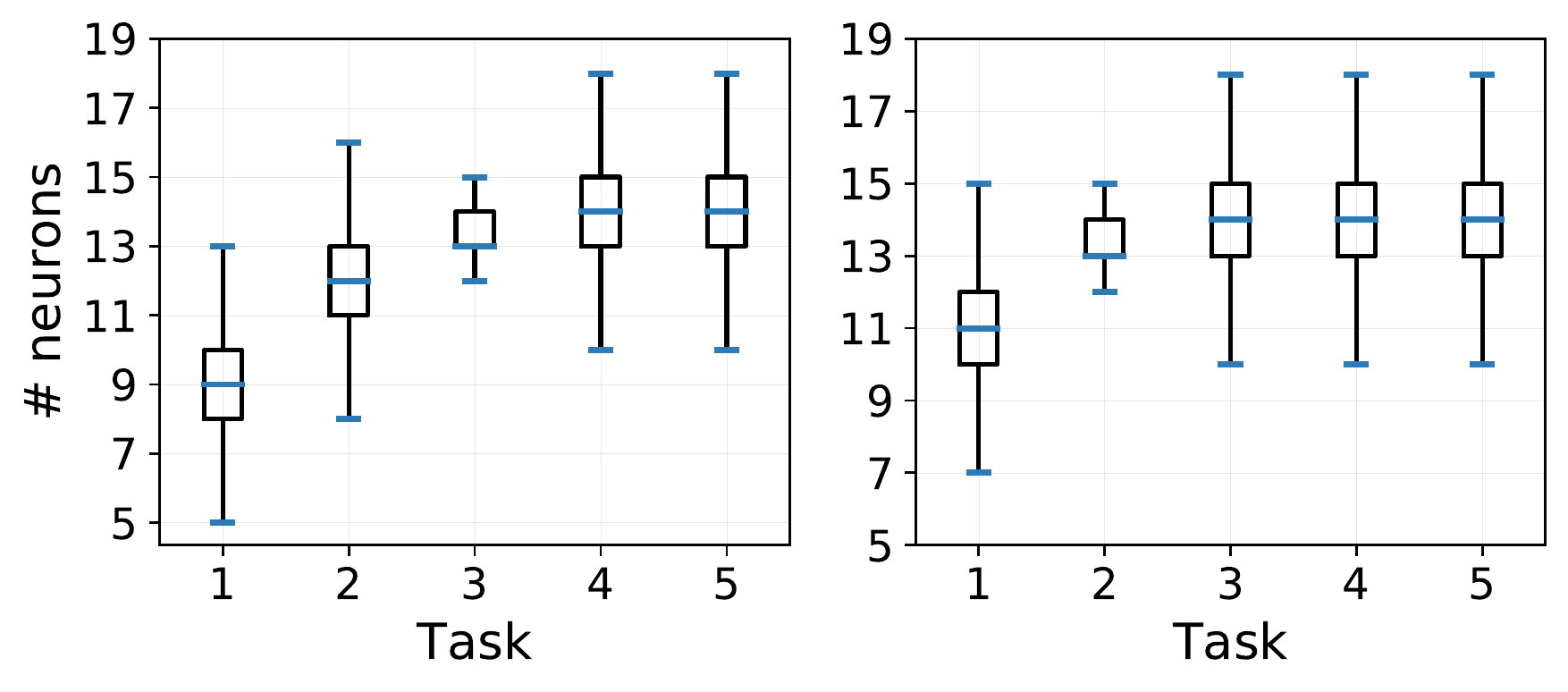}
     \caption{Box plots of the number of active neurons selected by the IBP variational posterior for each task, for \textbf{Left} Split MNIST and \textbf{Right} Split MNIST background noise variant for \texttt{CL1}. For both datasets our model expands its capacity over the course of CL.}
      \label{fig:Zs_bp_cl1_normal_random_mnist}
\end{figure}

For CL on split MNIST, the IBNN is able to outperform the 10 neuron VCL baseline as it underfits. On the other hand, the larger VCL networks slightly outperform the IBNN, see Figure~\ref{fig:accs_all_normal_mh}. The IBP prior enables the BNN to expand from a median of 11 neurons for the first task to 14 neurons for later tasks, see Figure~\ref{fig:Zs_bp_cl1_normal_random_mnist}. 

Regarding CL on MNIST with random background noise, it is clear that the IBNN is able to outperform VCL for all widths considered. The baseline models overfit on the second task and propagate a poor approximate posterior which affects subsequent task performance, see Figure~\ref{fig:accs_all_random_mh}. The IBNN expands over the course of the 5 tasks from a median of 11 neurons to 14 neurons in Figure~\ref{fig:Zs_bp_cl1_normal_random_mnist}.

\begin{figure*}
     \centering
     \begin{subfigure}{\textwidth}
        \centering
        \includegraphics[width=0.99\textwidth]{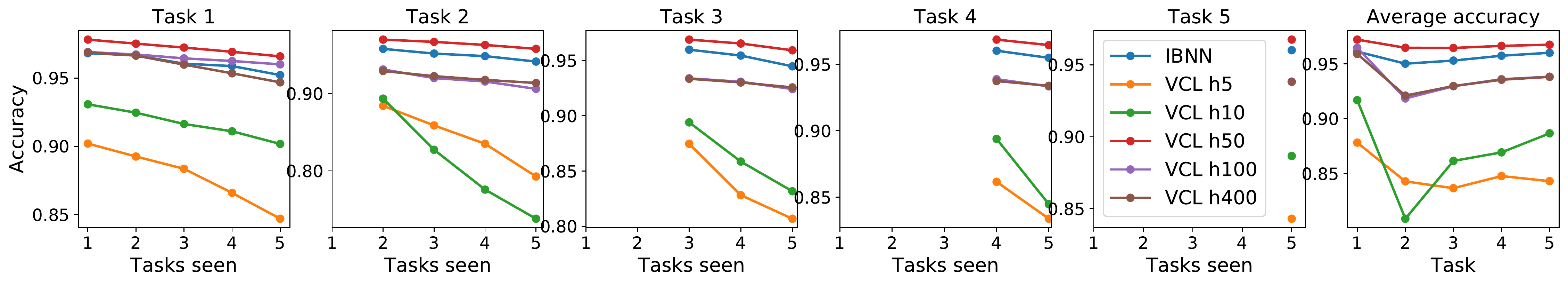}
     \end{subfigure}
     \caption{Permuted MNIST task accuracies versus the number of tasks the model has seen and performed a Bayesian update. Our model is compared to VCL with different numbers of hidden states sizes. Our model has a very good performance versus most models for Permuted MNIST.}
      \label{fig:accs_all_perm_mh}
\end{figure*}

\begin{figure*}
     \centering
     \begin{subfigure}{\textwidth}
        \centering
        \includegraphics[width=0.99\textwidth]{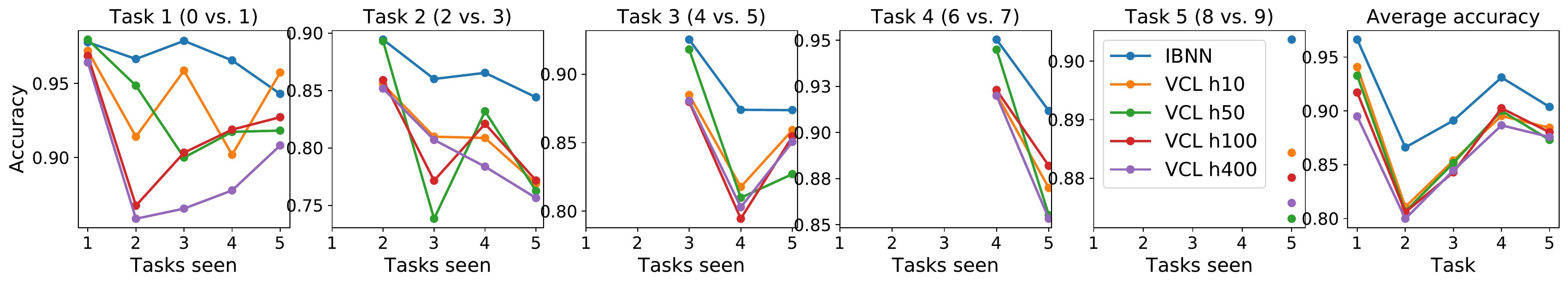}
     \end{subfigure}
     \caption{Split MNIST with background images, task accuracies versus the number of tasks the model has seen and performed a Bayesian update. Our model is compared to VCL with different hidden state sizes. Our model adapts its complexity overcomes forgetting better than VCL.}
      \label{fig:accs_all_split_images_mh}
\end{figure*}

VCL networks with a small width of $10$ neurons on Permuted MNIST display underfitting due to a lack of capacity which translates into forgetting in CL. On the over hand, a VCL model which is overparameterised with a width of $100$ displays overfitting and will subsequently propagation of a poor posterior and results in forgetting for all future tasks, Figure~\ref{fig:accs_all_perm_mh}. The same phenomena of forgetting due to underfitting due to restricted model size and overfitting due to overparameterisation is observed for the Split MNIST with random background images dataset in Figure~\ref{fig:accs_all_split_images_mh}.

\section{Experimental Details}
\label{sec:exp_details}
We summarise all dataset sizes in Table \ref{tab:data_sizes}.
\paragraph{Permuted MNIST.} No preprocessing of the data is performed like \cite{VCL}. For the permuted MNIST experiments, the BNNs used for the baselines and the IBNN consist of a single layer with ReLU activations. For the IBNN, the variational truncation parameter is set to $K=100$. For the first task the parameters of the Beta prior and the variational Beta distribution are initialised to $\alpha_k = 5$ and $\beta_k = 1$ for all $k$. The temperature parameters of the Concrete distributions for the variational posterior and priors are set to $\lambda_1 = 1$ and $\lambda_2 = 1$ respectively. For each batch, $10$ samples are averaged from the IBP priors as this made training stable. The Gaussian weights of the BNNs have their means and variances initialised with the maximum likelihood estimators found after training for 100 epochs with Adam and the variances initialised to $\log \sigma^2 = -6$ for the first task. For the CL Adam is also used 200 epochs and an initial learning rate of 0.001 which decays exponentially with a rate of 0.87 every 1000 iterations for each task. The CL experiments are followed the implementation of VCL \citep{VCL}.

\paragraph{Split MNIST and variants.}  The BNNs used for the baselines and the IBNN consist of a single layer with ReLU activations. The variational truncation parameter is set to $K=100$. For \texttt{CL1} and for the first task, the parameters of the Beta prior and the variational Beta distribution are initialised to $\alpha_k = 5$ and $\beta_k = 1$ for all $k$. The temperature parameters of the Concrete distributions for the variational posterior and priors are set to $\lambda_1 = 0.7$ and $\lambda_2 = 0.7$, respectively. The prior for the Gaussian weights is $\mathcal{N}(0, 0.7)$.  The Gaussian weights of the BNNs have their means initialised with the maximum likelihood estimators, found after training a NN for 100 epochs with Adam. The variances initialised to $\log \sigma^2 = -6$. The Adam optimiser is used to for $600$ epochs, the default in the original VCL paper \citep{VCL} for Split MNIST. To ensure convergence of the IBNN, the first task needs to be trained for $20\%$ more epochs. An initial learning rate of $0.001$ which decays exponentially with a rate of $0.87$ every $1000$ iterations is employed.

\subsection{Hyperparameter optimisation details}
\label{sec:rs}
Random search is performed on some of the experiments presented in the main paper by sampling over a discrete set of values or over a range for some hyperparameters listed below. We denote curly brackets $\{\ldots\}$ as a discrete set and square brackets $[ \ldots]$ as a range. 

HIBNN \texttt{CL1} and \texttt{CL2} for CL experiments with increasing task difficulty (Table~\ref{tab:exp_accs}).
\begin{itemize}
    \item Concrete posterior temperature: $\{1/2, 2/3, 3/4, 1, 5/4, 3/2, 7/4, 2, 9/4, 5/2, 11/4, 3\}$.
    \item Concrete prior temperature: $\{1/2, 2/3, 3/4, 1, 5/4, 3/2, 7/4, 2, 9/4, 5/2, 11/4, 3\}$.
    \item Child IBP base $\alpha$: $\{1, 2, 3, 4, 5\}$.
    \item Child IBP multiple of $\alpha$ after each new dataset seen in during CL: $\{1, 2, 3\}$
    \item Global IBP prior, $\alpha^0$: $[5, 25]$
\end{itemize}

IBNN \texttt{CL1} and \texttt{CL2} for CL experiments with increasing task difficulty (Table~\ref{tab:exp_accs}) and IBNN \texttt{CL3} for Split MNIST with background noise and \texttt{CL2} and \texttt{CL3} for Split MNIST with background images (Table~\ref{tab:mnist_accs}).
\begin{itemize}
    \item Concrete prior temperature: $\{1/2, 2/3, 3/4, 1, 5/4, 3/2, 7/4, 2, 9/4, 5/2, 11/4, 3\}$
    \item Concrete posterior temperature: $\{1/2, 2/3, 3/4, 1, 5/4, 3/2, 7/4, 2, 9/4, 5/2, 11/4, 3\}$
    \item IBP prior, $\alpha$: $[5, 25]$
\end{itemize}

\subsection{Baselines Implementations} 

The implementations for EWC, SI and GEM is from \cite{Hsu2018} and uses default hyperparameters. For DEN we use the implementation provided by the authors with default hyperparameters \citep{Yoon}. For VCL we likewise use the implementation provided by the authors with default hyperparameters \citep{VCL}.

\subsection{Training time}
\label{sec:training_times}
\begin{SCfigure}
     \centering
      \caption{Time to run one permuted MNIST task for our model and VCL. Our model takes longer due to inference of the IBP variational posterior.}
     \includegraphics[width=0.3\textwidth]{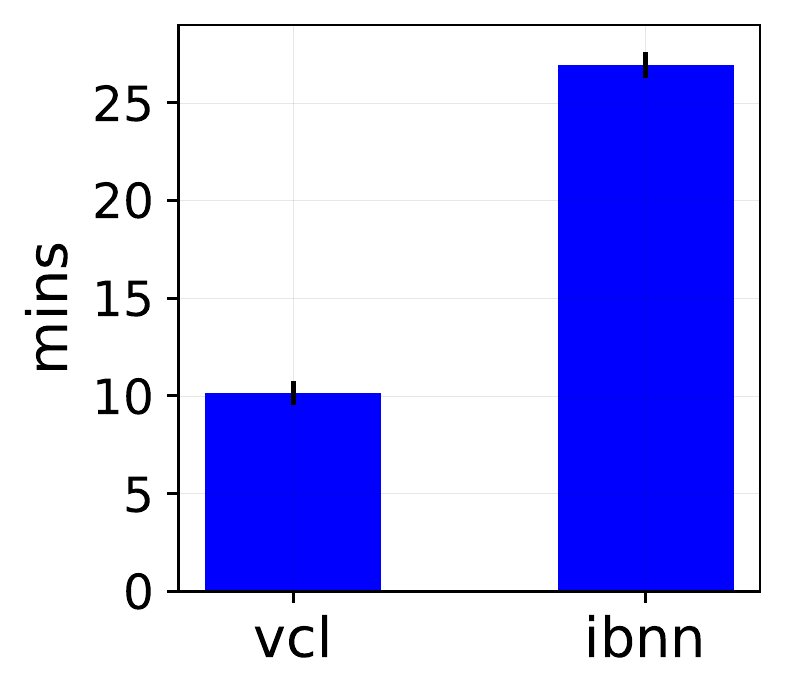}
      \label{fig:perm_timings}
\end{SCfigure}

Comparing the training time of one Permuted MNIST task for $200$ epochs can be seen in Figure~\ref{fig:perm_timings}. The IBNN takes longer to train. Stochastic variational inference of the IBP posterior involves taking multiple samples from it in forward pass. Then in the backward pass, gradients of samples from the Concrete distribution and implicit gradients from samples from the Beta distribution need to be calculated; this explains the longer training times in comparison to VCL. Improving the inference scheme is a good direction for further work.

\begin{table}
\centering
\caption{Dataset sizes used for experiments. Split MNIST's variants is obtained from \url{https://sites.google.com/a/lisa.iro.umontreal.ca/public\_static\_twiki/variations-on-the-mnist-digits}.}
\begin{tabular}{lll}
\toprule
 Dataset &  Train set size & Test set size \\
 \midrule
 Permuted MNIST & $50,000$ & $10,000$ \\
 Split MNIST & $60,000$ & $10,000$ \\
 Split MNIST + noise & $52,000$ & $10,000$ \\
 Split MNIST + images & $52,000$ & $10,000$ \\
 CIFAR 10 & $50,000$ & $10,000$ \\ 
 \bottomrule
\end{tabular}
\label{tab:data_sizes}
\end{table}

\section{Detailed Comparison with Related Works}
\label{sec:comparison}
The related papers \citep{Panousis2019} and \citep{Kumar} apply the IBP prior to a BNN like in our model, however they apply it in a very different fashion. In these works, the IBP prior is applied to each layer such that $Z \in \Z^{d \times k}_{2}$, where $d$ is input dimension or input hidden layer size and $k$ is the output dimension. Crucially, $Z$ is not sampled for each data point. Our model applies $Z \in \Z^{n \times k}_{2}$, such that each point selects neurons according to the IBP, this remains closer to the original formulation of the IBP prior for matrix factorisation, Section~\ref{sec:ibp} and to related works applying the IBP prior to VAEs.
\begin{figure}
     \centering
     \begin{subfigure}{0.24\textwidth}
        \includegraphics[width=1\textwidth]{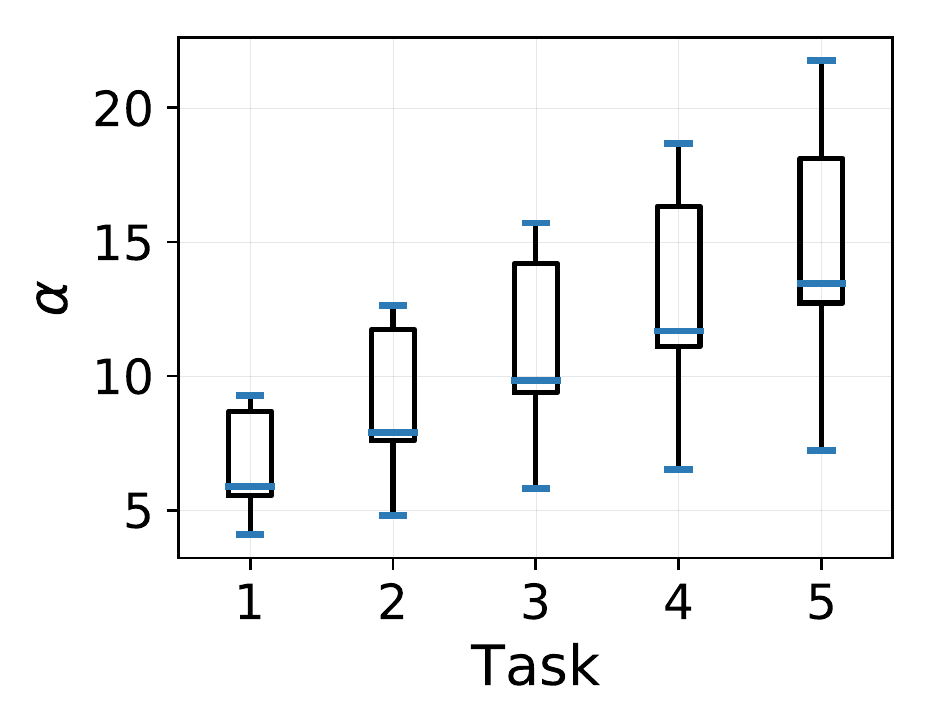}
        \caption{IBNN $\alpha$.}
        \label{fig:perm_alphas_cl1_ibp}
     \end{subfigure}
     \hfill
     \begin{subfigure}{0.24\textwidth}
        \includegraphics[width=1\textwidth]{pics/Zs_bp_perm_l1.pdf}
        \caption{IBNN $Z$.}
        \label{fig:perm_Zs_cl1_ibp}
     \end{subfigure}
     \hfill
     \begin{subfigure}{0.24\textwidth}
        \includegraphics[width=1\textwidth]{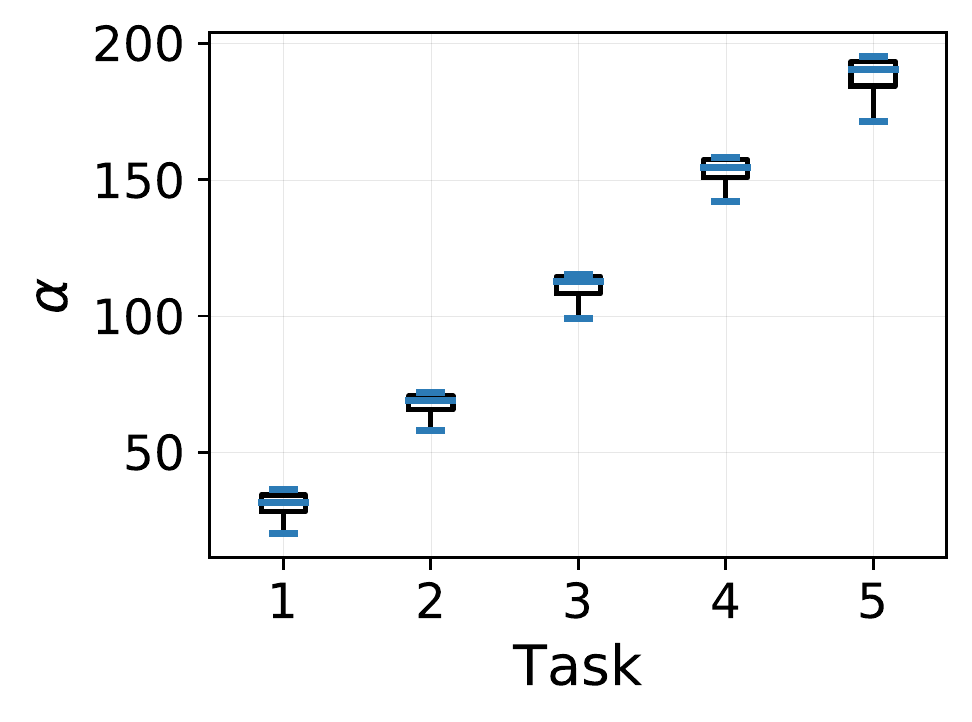}
        \caption{BSCL $\alpha$.}
        \label{fig:perm_alphas_bscl}
     \end{subfigure}
     \hfill
     \begin{subfigure}{0.24\textwidth}
        \includegraphics[width=1\textwidth]{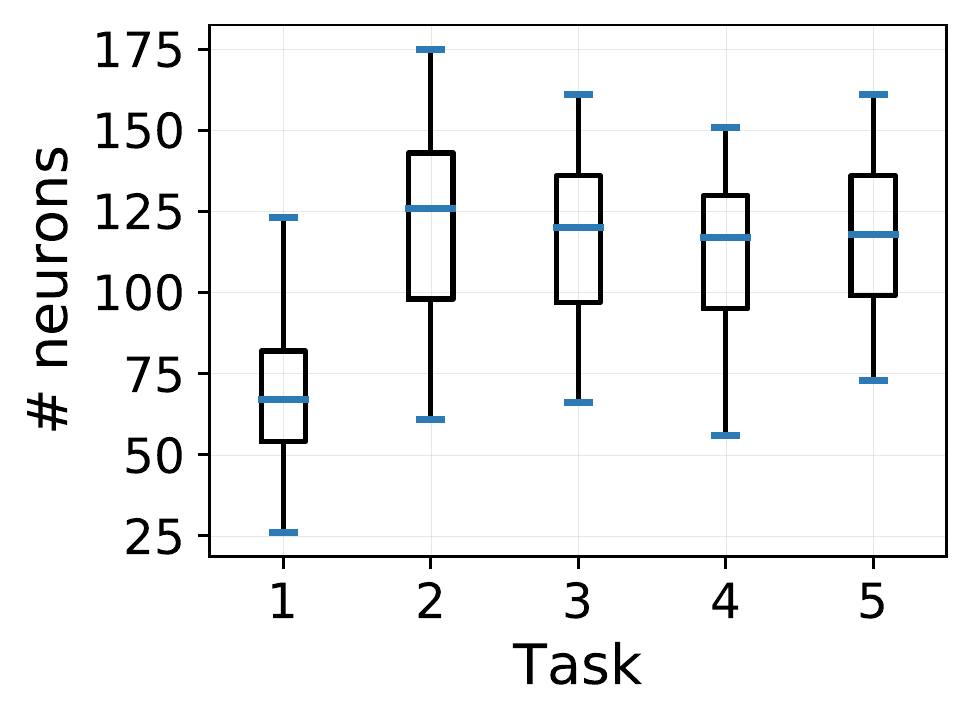}
        \caption{BSCL $Z$.}
        \label{fig:perm_Zs_bscl}
     \end{subfigure}
     \caption{Comparing our model to BSCL \citep{Kumar} for a simple $5$ task CL experiment on Permuted MNIST. In our model the variational parameters $\alpha$ which control the number of neurons, increases for each task, Figure~\ref{fig:perm_alphas_cl1_ibp}, as does the resulting number of neurons, Figure~\ref{fig:perm_Zs_cl1_ibp}. While for BSCL as $\alpha$ increases, Figure~\ref{fig:perm_alphas_bscl}, this doesn't translate into additional weights becoming activated, Figure~\ref{fig:perm_Zs_bscl}. Since the additional parameter $p_k$ for all $k$ is influencing the number of neurons which are active in addition to the IBP prior term item~\ref{item:1}. We use the outputs of the Concrete distribution as $Z$ and define a neuron as active if $z_{k} > 0.1$ for neuron $k$.}
      \label{fig:kumar_comp}
\end{figure}

\begin{figure}
     \centering
     \includegraphics[width=0.5\textwidth]{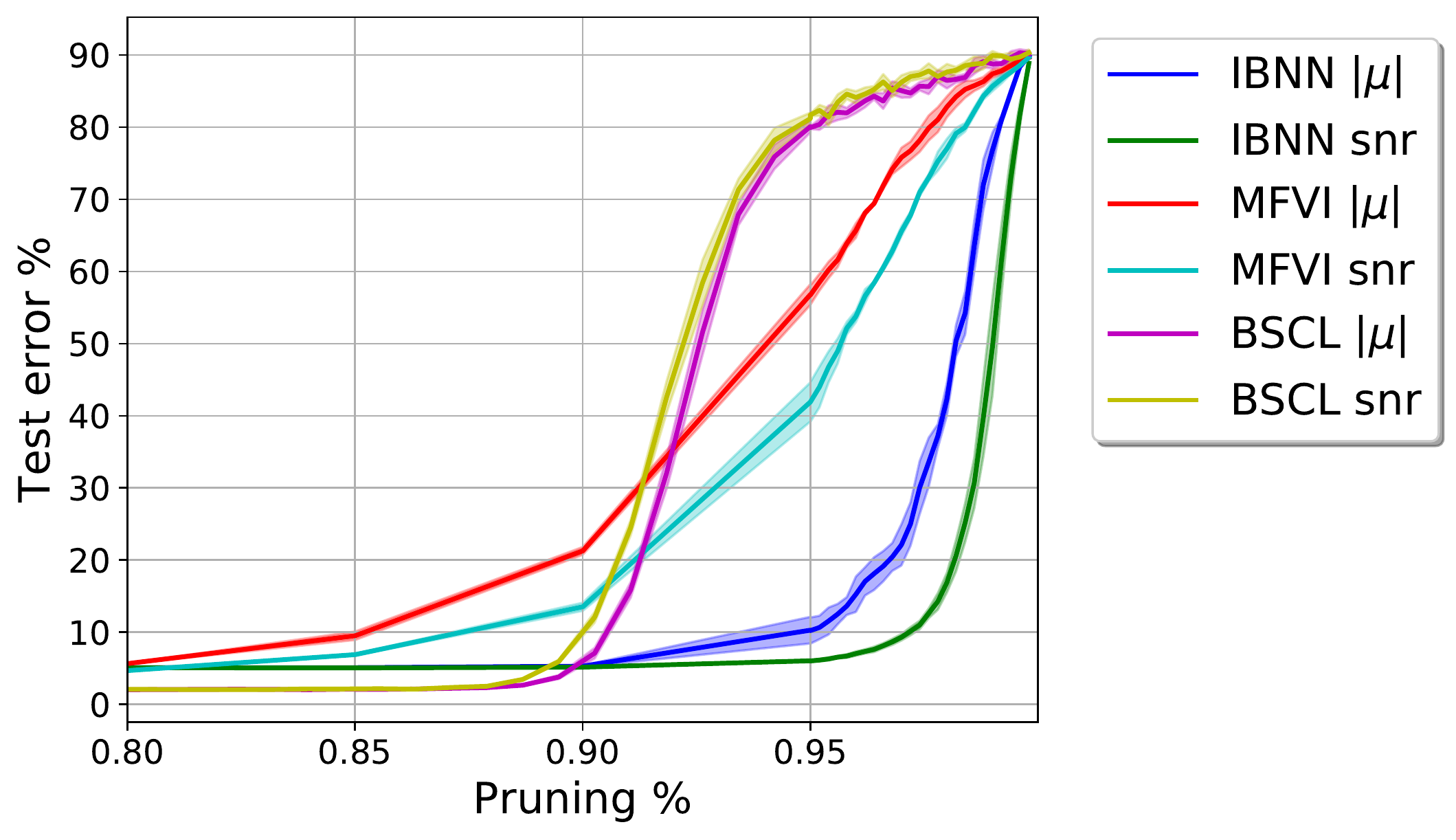}
     \caption{Weight pruning on MNIST for BSCL compared to our model and a BNN with independent Gaussian weights (MFVI). BSCL is less robust to pruning than our model and MFVI, also it is less robust when pruning with signal to noise ratio: $|\mu|/\sigma$, thus the variational variances are not being learnt properly.}
      \label{fig:wp_bscl}
\end{figure}

Despite this important difference with Bayesian Structure Adaptation for CL (BSCL) \citep{Kumar}, the results for BSCL are strong and suggest some interesting design choices and ideas that can be leveraged to use the IBP prior for CL with BNNs. It is promising to see similar ideas being put forward in this area. There are some design choices which are quite different from ours in three main regards.
\begin{enumerate}
    \item \label{item:1} An additional variational parameter for each Gaussian neural network weight is introduced. The IBP prior model is $z_{k}\sim \textrm{Ber}(\pi_k)$ where $\pi_k = \prod^{k}_{i=1} v_{i}$ and where $v_i \sim \textrm{Beta}(\alpha, \beta)$ for each layer, for a neuron $k$. However in the BSCL implementation, the Bernoulli probability is $\tilde{\pi}_k = p_k + \prod^{k}_{i=1}v_i$, where $p_k$ is an additional learnable parameter. Thus the Bernoulli parameters are being directly learnt, similarly to \cite{Dikov2019}. The parameter $\alpha$ directly controls the number of active items in the IBP prior \citep{Griffiths2011} and so this additional parameter $p_k$ can potentially dominate $\alpha$ in controlling the expansion of $Z$ see Figure~\ref{fig:kumar_comp}. 
    \item \label{item:2} The implementation of BSCL does not strictly follow sequential Bayes, in particular the Beta distribution (which is part of the IBP and controls expansion). In sequential Bayes, like VCL, the variational posterior of the Beta distribution (the implementation uses a relaxation, the Kumaraswamy distribution), is used as the prior for the forthcoming task. The previous task's variational Beta posterior parameters are $\alpha \in \R_{+}^K$ and $\beta \in \R_{+}^{K}$ respectively, the new task's Beta prior parameters are then set to $\textrm{max}(\alpha) \in \R_{+}$ and $1$ respectively where $K$ is the variational truncation\footnote{\url{https://github.com/scakc/NPBCL/blob/master/ibpbnn.py\#L939}}. This explains the large increase in the values of $\alpha$ seen in Figure~\ref{fig:kumar_comp} and so allows a lot of expansion in the model.
    \item BSCL requires more memory in storing $Z$'s from each task. The $Z$'s are used as a mask to then finetune the network which can result in better performance. For each task, $Z$'s for all weights are stored and recalled thus alleviating forgetting, rather than relying on sequential variational Bayes like VCL and our method. We apply sequential Bayes on the IBP and so generate $Z$'s for each new task without having to store any parameters from previous tasks.
\end{enumerate}
\begin{table}
  \centering
  \caption{Average test accuracy on MNIST variants for \textit{task incremental learning} (\texttt{CL1}) experiments over 5 runs. Despite the many design choices BSCL \citep{Kumar}, performs similarly to our method which is simpler and more correct.}
\begin{tabular}{lll}
\toprule
  & BSCL & IBNN \\
 \midrule
 S+$\epsilon$ \texttt{CL1} & $\bf{95.2}$\scalebox{0.85}{$\bf{\pm 1.5}$} &$\bf{95.1}$\scalebox{0.85}{$\bf{\pm 1.1}$} \\
 S+img \texttt{CL1} & $\bf{92.9}$\scalebox{0.85}{$\bf{\pm 1.0}$} & $91.6$\scalebox{0.85}{$\pm 1.2$} \\
 \bottomrule
\end{tabular}
  \label{tab:mnist_accs_bscl}
\end{table}
Despite these design choices, BSCL performs similarly to our model for the more difficult MNIST variants, see Table~\ref{tab:mnist_accs_bscl}. We use the implementation provided by the authors with default parameters used for Split MNIST CL experiments. We also perform weight pruning on MNIST using a $2$ layer BSCL network with variational truncation $K=200$ using the default parameters for Split MNIST CL experiments. We see from Figure~\ref{fig:wp_bscl} that despite having a strong $0\%$ pruning accuracy of $98\pm 0.1$ in comparison to the IBNN's $95\pm 0.0$. As we prune weights by the signal to noise ratio (snr), $|\mu| / \sigma$, this is less robust than pruning by $|\mu|$ which shows that BSCL has not learnt proper variational variances. 

\end{document}